\documentclass{article}

\usepackage{amsmath,amsfonts,bm}

\def\eqref#1{equation~\ref{#1}}

\def\1{\bm{1}}

\def\rva{{\mathbf{a}}}
\def\rvb{{\mathbf{b}}}

\def\rvu{{\mathbf{i}}}

\def\rvr{{\mathbf{r}}}

\def\rvu{{\mathbf{u}}}

\def\rvw{{\mathbf{w}}}
\def\rvx{{\mathbf{x}}}
\def\rvy{{\mathbf{y}}}
\def\rvz{{\mathbf{z}}}

\def\rmW{{\mathbf{W}}}

\DeclareMathAlphabet{\mathsfit}{\encodingdefault}{\sfdefault}{m}{sl}
\SetMathAlphabet{\mathsfit}{bold}{\encodingdefault}{\sfdefault}{bx}{n}

\usepackage{microtype}
\usepackage{graphicx}
\usepackage{subfigure}
\usepackage{booktabs} 
\usepackage{makecell}
\usepackage{ dsfont }
\usepackage{pdflscape}
\usepackage{afterpage}
\usepackage{ textcomp }
\usepackage{multirow}
\usepackage{ wasysym }
\usepackage{xcolor}
\usepackage{hyperref}
\hypersetup{colorlinks=true}

\usepackage[preprint,nonatbib]{neurips_2021}

\usepackage[utf8]{inputenc}
\usepackage{amsmath}
\usepackage{amssymb}
\usepackage{mathrsfs}

\newcommand{\indep}{\perp \!\!\! \perp}

\title{{Integrating Expert ODEs into Neural ODEs: \\
Pharmacology and Disease Progression}}

\author{%
  Zhaozhi Qian \\
  University of Cambridge \\
  \texttt{zhaozhi.qian@maths.cam.ac.uk} \\
   \And
   William R. Zame \\
   UCLA \\
   \texttt{zame@econ.ucla.edu} \\
   \And
   Lucas M. Fleuren \\
   Amsterdam UMC \\
   \texttt{l.fleuren@amsterdamumc.nl} \\
   \And
   Paul Elbers \\
   Amsterdam UMC \\
   \texttt{p.elbers@amsterdamumc.nl} \\
    \AND
   Mihaela van der Schaar \\
   University of Cambridge \\
   UCLA \\
   The Alan Turing Institute \\
   \texttt{mv472@cam.ac.uk} \\
}

\begin{document}

\maketitle

\begin{abstract}
Modeling a system's temporal behaviour in reaction to external stimuli is a fundamental problem in many areas.  Pure Machine Learning (ML) approaches often fail in the small sample regime and cannot provide actionable insights beyond predictions.  A promising modification has been to incorporate expert domain knowledge into ML models. 
The application we consider is predicting the progression of disease under medications, where a plethora of domain knowledge is available from pharmacology. 
Pharmacological models describe the dynamics of carefully-chosen medically meaningful variables in terms of systems of Ordinary Differential Equations (ODEs).  
However, these models only describe a limited collection of variables, and these variables are often not observable in clinical environments. 
To close this gap, we propose the latent hybridisation model (LHM) that integrates a system of expert-designed ODEs with machine-learned Neural ODEs to fully describe the dynamics of the system and to link the expert and latent variables to observable quantities.
We evaluated LHM on synthetic data as well as real-world intensive care data of COVID-19 patients.
LHM consistently outperforms previous works, especially when few training samples are available such as at the beginning of the pandemic.
\end{abstract}

\section{Introduction}  
{
Understanding the evolution of a system in response to stimuli is  {\em the} central problem in many areas. The Machine Learning (ML) approach to this problem has been to learn a collection of latent variables and construct a dynamical model of the system directly from observational data.
While ML has achieved strong predictive performance in some applications, it has two central weaknesses. 
The first is that it requires large datasets.  The second is that the latent variables that the ML approach identifies often have no physical interpretation and do not correspond to any previously-identified quantities. }

{
One approach to dealing with these weaknesses has been to incorporate expert domain knowledge into  ML models.  
Most  of the work using this approach has focused on incorporating \textit{high-level} knowledge about the underlying physical system, such as conservation of energy \cite{bertalan2019learning,greydanus2019hamiltonian,zhong2019symplectic}, independence of mechanism \cite{parascandolo2018learning}, monotonicity \cite{muralidhar2018incorporating}, or linearity \cite{guen2020disentangling}.  In addition, there have been attempts to  integrate domain-specific ``expert models'' into ML models to create ``hybrid'' models.  Most of this work has employed expert models that  directly issue predictions \cite{liu2019multi,wang2017physics,xu2015data,yao2018tensormol} or  extract useful features from the raw measurements \cite{karpatne2017physics}. }

{
The approach taken in this paper begins with an expert model in the form of a system of  Ordinary Differential Equations (ODEs) and integrates that expert model into a system of Neural ODEs \cite{chen2018neural}.  The specific problem we address is that of predicting disease progression under medications;  the specific expert model(s) come from Pharmacology \cite{katzung2012basic} -- but we think our approach may be much more widely applicable.} For a number of diseases, available pharmacological models, built on the basis of specialized knowledge and laboratory experiments, provide a  description of the dynamics of carefully-chosen medically meaningful variables in terms of ODEs that govern the evolution of these states \cite{danhof2007mechanism,agoram2007role,gesztelyi2012hill}.
However, these models are typically not directly applicable in clinical environments, because they involve too few variables to fully describe a patient's health state \cite{spoorenberg2014pharmacokinetics,holford2013pharmacokinetic}, because the expert variables which the models employ may be observable in the laboratory setting but not in clinical environments \cite{frank2003clinical, aronson2017biomarkers}, and because the relationships between the expert variables and clinically observable quantities is not known \cite{falvey2015disease}. We will give a  example later in Section \ref{sec:decision}.

This paper proposes a novel hybrid modeling framework, the Latent Hybridisation Model (LHM), that  imbeds a given pharmacological model (a collection of expert variables and the  ODEs that describe the evolution of these variables ) into a larger latent variable ML model (a system of Neural ODEs).  In the larger model, we use observational data to learn {\em both} the evolution of the unobservable latent variables and the relationship between measurements and {\em all} the latent variables --  the expert variables from the pharmacological model {\em and} the latent variables in the larger model.   The machine learning component provides links between the expert variables and the clinical measurements, the underlying pharmacological model improves sample efficiency, and the expert variables provide additional insights to the clinicians.  A variety of experiments (using synthetic and real data) demonstrate the effectiveness of our hybrid approach.

\section{Problem setting}

We consider a set of hospitalized patients $[N] = \{1, \ldots, N\}$ over a time horizon $[0,T]$; $t=0$ represents the time of admission and $t=T$ represents the maximal length of stay.  
The health status of each patient $i$ is characterized by a collection of  \textit{observable} physiological variables $\rvx_i(t) \in \mathbb{R}^D$, $D\in\mathbb{N}^+$; because the physiological variables may include vital signs, bloodwork values, biomarkers, etc., $\rvx_i(t)$ is typically a high-dimensional vector. Although the physiological variables are observable, they are typically \textit{measured} only at discrete times, and with error.
To avoid confusion, we distinguish the measurements of these variables $\rvy(t)$ from the true values; i.e. 
\begin{equation}
\label{eq:error}
\rvy_i(t) = \rvx_i(t) + \epsilon_{it}
\end{equation}
where the independent noise term $\epsilon_{it}$ accommodates the measurement error (modeling $\epsilon_t$ as an autocorrelated stochastic process is left as a future work).
For illustrative purposes, we also assume that  $\epsilon_t$ follows a Normal distribution $N(0, \sigma^2_i)$, but any parametric distribution could be easily accommodated.
We denote the measurement times for each patient as $\mathcal{T}_i = \{t_{i1}, t_{i2}, \ldots\}$. 
We write $\rva_i(t) \in \mathbb{R}^A$, $A\in\mathbb{N}^+$ for the {\em treatments} the patient receives.  Some treatments (e.g. intravenous medications) are continuous; others (e.g. surgical interventions) are discrete, so some components of $\rva_i(t)$ may be continuous functions but others are (discontinuous) step functions.  

It is convenient to write $\mathcal{A}_i[t_1:t_2] = \{\rva_i(t) | t_1 \le t \le t_2\}$ and $\mathcal{Y}_i[t_1:t_2]  = \{\rvy_i(t) | t_1 \le t \le t_2, t \in \mathcal{T}_i \}$ for the treatments and measurements (respectively) during the the time window $[t_1,t_2]$.  Note that $\mathcal{Y}_i[0:t]$ and 
$\mathcal{A}_i[0:t]$ represent {\em histories} at time $t$ while $\mathcal{A}_i[t:T]$ and $\mathcal{Y}_i[t:T]$ represent {\em treatment plans} and {\em predictions}, respectively. 
Our objective is to 
{\em predict }the future measurements under a given treatment plan ${\mathcal{A}}_i[t_0:T]$ given the history:
\begin{equation}
\label{eq:1}
\mathbb{P}(\mathcal{Y}_i[t_0:T]\ |\ \underbrace{\mathcal{Y}_i[0:t_0], \mathcal{A}_i[0:t_0]}_\text{Historical observations}, \underbrace{{\mathcal{A}}_i[t_0:T]}_\text{Treatment plan}).
\end{equation}
Understanding this distribution will allow us to compute both point estimates and credible intervals (reflecting uncertainty).  Note that it is important for the clinician to understand uncertainty in order to  balance risk and reward. When the context is clear, we will omit the subscript $i$ and the time index $t$.

\section{Method}

\begin{figure}[t!]
  \centering
  \includegraphics{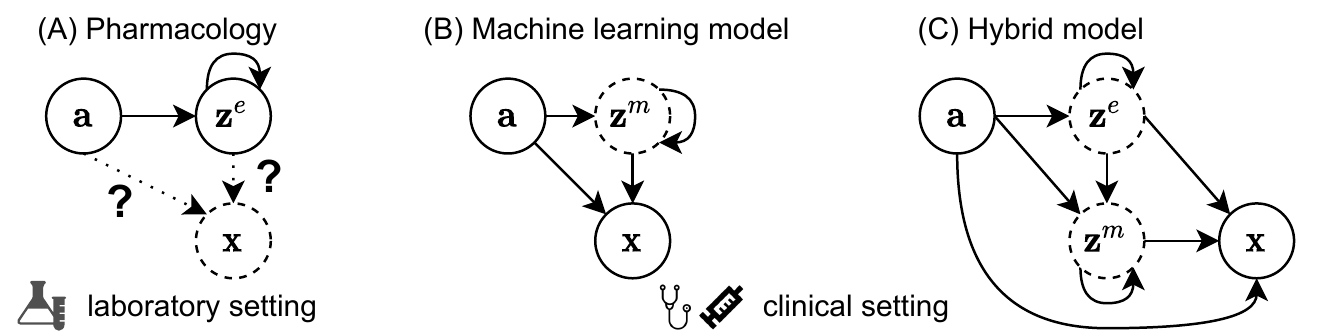}
  \caption{\footnotesize Dependency structure of the three models designed for the laboratory or clinical settings. Dashed nodes represent unobservable variables. The expert variables $\rvz^e$ are observable in the laboratory setting but not in the clinical setting. The pharmacological model does not contain the links to the clinical variables $\rvx$.}
  \label{fig:1}
\end{figure}

\subsection{The pharmacological model}
We begin with a  pharmacological model which describes the  dynamics of a collection of ``expert'' variables  $\rvz^e(t) \in \mathbb{R}^E$. Each expert variable captures a distinct and medically-meaningful aspect of the human body, e.g. the activation of immune system. 
The pharmacological model describes the dynamics as a system of Ordinary Differential Equations (ODEs):
\begin{equation}
\label{eq:expert}
\dot{\rvz}^e(t) = f^e(\rvz^e(t), \rva(t); \theta^e),
\end{equation}
where we have written $\dot{\rvz}^e(t) $ for the time derivative of $\rvz^e$.   The functional form of 
$f^e: \mathbb{R}^E \times \mathbb{R}^A \rightarrow \mathbb{R}^E$  is specified but the  unknown parameters $\theta^e$ (e.g., coefficients) need to be estimated from data.\footnote{The system in Equation (\ref{eq:expert}) is quite general; appropriate choices of expert variables allow it to  capture both high-order ODEs and time-dependent ODEs \cite{perko2013differential}.}${}^,$\footnote{Some care must be taken because  systems such as (\ref{eq:expert}) do not always  admit unique global solutions.  In practice, the pharmacological models are sufficiently well-behaved that  global solutions  exist and are  unique. Although closed-form solutions may not be available, there are various efficient numerical methods for solution.}  

It is important to note that the system of ODEs (\ref{eq:expert}) describes dynamics that are \textit{self-contained}, in the sense that the time derivatives $\dot{\rvz}^e(t)$ depend only on the {\em current} values of the expert variables $\rvz^e(t)$ and the {\em current}  treatments 
$\rva(t)$, and not on histories or on other variables.  To ensure that this obtains, it may be necessary to limit the scope of the model and limit attention to a single system of the body (or perhaps to several closely related systems) \cite{danhof2016systems}.   As a consequence of these limitations, the expert variables will usually not give a full picture of the health status of the patient and will usually not account for the full array of {\em observable} physiological variables  $\rvx(t)$ \cite{frank2003clinical}.

\subsection{The latent hybridisation model: linking expert variables with measurements}

As we have already noted,  pharmacological models are typically developed and calibrated in the laboratory, where the expert variables can be directly measured -- in patients, in laboratory animals, or even in vitro (Figure \ref{fig:1} A).  In clinical environments, the expert variables are frequently not observed (Figure \ref{fig:1} B and C).  To use the pharmacological models in  clinical environments, we must establish links between the expert variables $\rvz^e(t)$ and the clinical measurements $\rvy(t)$.  To do this we introduce additional latent variables  
$\rvz^m(t)\in \mathbb{R}^M$ and posit the following relationship between the latent variables $\rvz^e, \rvz^m$ and the observable physiological variables  $\rvx(t)$:
\begin{equation}
 \label{eq:physiological}
\rvx(t) = g(\rvz^e(t), \rvz^m(t), \rva(t);\ \gamma)
\end{equation}
The function $g: \mathbb{R}^{E \times M \times A} \rightarrow \mathbb{R}^D$ is a neural network with (unknown) weights $\gamma$, and maps the latent space to the ``physiological space''.   We also posit that the dynamics of the latent variables 
$\rvz^m(t)$ follow a system of ODEs governed by its current values ${\rvz}^m(t)$, the treatments $\rva(t)$ {\em and} the current values of the expert variables $\rvz^e(t)$. 
\begin{equation}
\label{eq:latent}
    \dot{\rvz}^m(t) = f^m(\rvz^m(t), \rvz^e(t), \rva(t);\ \theta^m),
\end{equation}
The function $f^m: \mathbb{R}^{M \times E \times A} \rightarrow \mathbb{R}^M$ is a neural network with (unknown) weights $\theta^m$. 
Equations (\ref{eq:expert})-(\ref{eq:latent}) specify the dynamics of LHM.  It is convenient to write $\rvz(t) = [\rvz^e(t)\ \rvz^m(t)]$ for the vector of all latent variables and $\Theta = (\theta^e, \theta^m, \gamma, \sigma)$ for the set of all (unknown) coefficients.

The coefficients $\Theta$ will be learned from data.  However, even after these coefficients are learned, the initial state of the patient $\rvz_i(0)$ is still {\em unknown}.  
In fact, the variation in initial states reflects the heterogeneity of the patient population. 
If the coefficients and the initial state were known, the entire trajectory of $\rvz_i$ given the treatments could be computed (numerically).  Because we have assumed that the noise/errors $\epsilon_t$ are independent, we have
\begin{equation}
\label{eq:independence}
\mathcal{Y}_i[t_0:T] \indep \mathcal{Y}_i[0:t_0]\ |\ \rvz_i(0), \mathcal{A}_i[0:T], \Theta, \quad \forall t_0 < T
\end{equation}
However, because the initial state is unknown, it must be {\em learned} from the measurements $\rvy_i(t)$. 

LHM would reduce to a pure latent neural ODE model  \cite{chen2018neural,rubanova2019latent} if we omitted the expert variables (Figure \ref{fig:1} B).  However, that would amount to {\em discarding prior (expert) information} and so is evidently undesirable.  Indeed, as we have noted in the introduction, our approach is driven by the idea of incorporating this prior (expert) information into our hybrid model.  

In the current work, we assume that the pharmacological model in Equation \ref{eq:expert} is correct. In practice, the model might be wrong in two ways.  The obvious way is that the functional form of $f^e$ might be misspecified (e.g. a linear model might be specified when the truth is actually nonlinear).  Many existing techniques can address such misspecification and could be integrated into LHM \cite{hamilton2017hybrid,parish2016paradigm,zhang2018real}; see the discussion in Appendix \ref{app:extensions}. Alternatively, it might be that the system of expert variables is {\em not} self-contained, and that their evolution actually depends on additional \textit{latent} variables,  we leave this more challenging problem for future work. 

Practical extensions to LHM such as including static covariates and modeling informative sampling are discussed in Appendix \ref{app:extensions}.

\subsection{Independent and informative priors}
It may be challenging to pinpoint the exact value of the latent variables $\rvz^e$ based on observations (e.g. due to measurement noise or sampling irregularity). 
For this reason, we quantify the uncertainty around $\rvz^e$  using Bayesian inference. 
In what follows, we assume the initial states $\rvz_i(0)$ of patients are independently sampled from a prior distribution $\rvz_i(0) \sim \mathbb{P}_0$.  Two points are worth noting.  

\textbf{Independent Priors.}   We use  {\em independent} prior distributions on the expert variables $\rvz^e$ and the latent variables  $\rvz^m$, i.e.
    $\mathbb{P}(\rvz(0)) = \mathbb{P}(\rvz^e(0)) \times \mathbb{P}(\rvz^m(0))$.
This guarantees that information in  $\rvz^{m}(0)$ does not duplicate (any of the) information in $\rvz^{e}(0)$, which captures our belief that the latent variables are incremental to the expert variables. 
In addition, independent priors are also commonly used in Bayesian latent variable models such as variational autoencoders (VAEs) \cite{kingma2013auto, higgins2016beta}.

\textbf{Informative Priors } The prior distribution on the expert variables $\mathbb{P}(\rvz^e(0))$ should reflect domain knowledge. Such knowledge is usually available from previous studies in Pharmacology \cite{holford2013pharmacokinetic}. Using an informative prior tends to improve the estimation of latent variables, especially in small-sample settings \cite{lee2018determining}. Moreover, the expert variables usually take values in specific ranges (e.g. $[0, 10]$ \cite{katzung2012basic}) and going beyond the valid range may lead to divergence. The informative prior can encode such prior knowledge to stabilize training.

\begin{figure}[t!]
  \centering
  \includegraphics[width=0.8\textwidth]{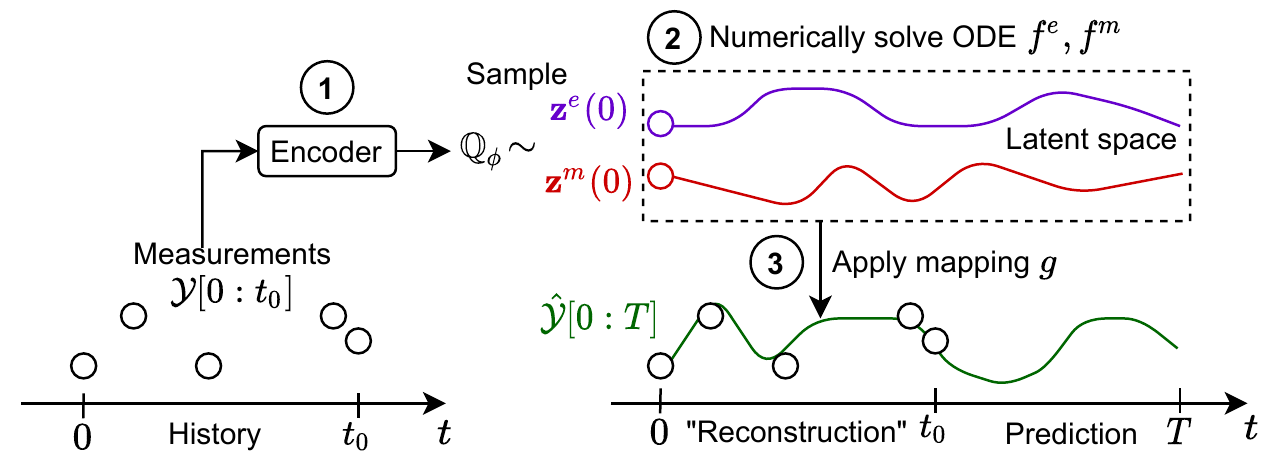}
  \caption{Illustration of the training and prediction procedure.}
  \label{fig:inference}
\end{figure}

\subsection{Model training and prediction via amortized variational inference}
\label{sec:inference}

Given the training dataset $\mathcal{D} = \{(\mathcal{Y}_i[0:T], \mathcal{A}_i[0:T])\}_{i \in [N]}$, we use amortized variational inference (AVI) to estimate the global parameters $\Theta$ and the unknown initial condition $\rvz_i(0)$ \cite{zhang2018advances}.  
Figure \ref{fig:inference} presents a diagram of the training procedure. We start by learning a \textit{variational distribution} to approximate the posterior $\mathbb{P}(\rvz_i(0)|\mathcal{Y}_i[0:T], \mathcal{A}_i[0:T])$. As is standard in AVI \cite{kingma2013auto, zhang2018advances}, 
we use a Normal distribution with diagonal covariance matrix as approximation:
\begin{equation}
\label{eq:variational}
 \mathbb{Q}(\rvz_i(0)|\mathcal{Y}_i[0:T], \mathcal{A}_i[0:T]) = N(\mu_i, \Sigma_i); \quad \mu_i, \Sigma_i = e(\mathcal{Y}_i[0:T], \mathcal{A}_i[0:T];\ \phi)
\end{equation}
Here the parameters $\mu_i, \Sigma_i$ are produced by an inference network (also known as an encoder) $e(\cdot)$ with trainable weights $\phi$. 
When the context is clear, we will denote the variational distribution defined by Equations (\ref{eq:variational}) as $\mathbb{Q}_\phi$.
The evidence lower bound (ELBO) for the global parameter $\Theta$ and the inference network parameters $\phi$ is defined as
\begin{equation}
 \text{ELBO}(\Theta, \phi) = \mathbb{E}_{\rvz(0) \sim \mathbb{Q}_\phi}\big[\text{log}\mathbb{P}(\mathcal{Y}_i[0:T] | \mathcal{A}_i[0:T], \rvz(0), \Theta)\big] - \text{KL}[\mathbb{Q}_\phi | \mathbb{P}_0]
\end{equation}
To compute the ELBO for a given $\Theta$ and $\phi$, we sample $\rvz(0) \sim \mathbb{Q}_\phi$ and numerically solve the ODEs to obtain $\rvz(t)$, $\forall t \in [0, T]$ (Figure \ref{fig:inference}; Steps 1, 2). Then, we compute the inner log-likelihood function using the mapping $g$ and the noise distribution (Equations (\ref{eq:error}) and (\ref{eq:physiological})). 
Finally, we use Monte Carlo sampling to evaluate the KL divergence term: $\mathbb{E}_{\rvz(0) \sim \mathbb{Q}_\phi}[\text{log}\mathbb{Q}_\phi(\rvz(0)) - \text{log}\mathbb{P}_0(\rvz(0))]$. This is because the informative prior $\mathbb{P}_0$ may not have an analytical KL divergence (unlike the standard Normal prior used in previous works \cite{kingma2013auto,rubanova2019latent}). 
We optimize ELBO by stochastic gradient ascent and update all parameters jointly in an end-to-end manner (detailed in Appendix \ref{app:optimization}). 

The prediction procedure follows the same steps as illustrated in Figure \ref{fig:inference}.
For a new patient with history $\mathcal{Y}_i[0:t_0], \mathcal{A}_i[0:t_0]$, we first estimate the variational posterior  $\mathbb{Q}_\phi$ using the trained encoder. From Equation (\ref{eq:independence}), we can estimate the target distribution in Equation (\ref{eq:1}) as:
\begin{equation}
    \mathbb{E}_{\rvz(0)\sim \mathbb{Q}_\phi}\big[\mathbb{P}\big(\rvy(t) | \rvz(0), \mathcal{A}_i[0:t_0], {\mathcal{A}}_i[t_0:T]\big)\big], \  \forall t > t_0.
\end{equation}
where ${\mathcal{A}}_i[t_0:T]$ is a future treatment plan.
The outer expectation can be approximated by Monte Carlo sampling from $\mathbb{Q}_\phi$ and the inner probability is given by the likelihood function. 

\textbf{Choice of variational distribution and encoder}. The training procedure above is agnostic to the exact choice of variational distribution and encoder architecture. We choose the Normal distribution to make fair comparisons with the previous works  \cite{chen2018neural,rubanova2019latent} . 
For the same reason, we use the reversed time-aware LSTM encoder proposed in \cite{chen2018neural}.
In the Appendix \ref{app:simulation}, we show additional experiments with more complex variational distributions, i.e. Normalizing Flows \cite{rezende2015variational}. 

\subsection{Using LHM to provide clinical decision support}
\label{sec:decision}

In order for clinicians to properly treat patients, they need to predict the progression of disease given the treatments.  Although machine learning models may demonstrate feature importance \cite{choi2016retain, alaaattentive}, they do not uncover the relationships between those features and the underlying pathophysiology. LHM can provide the missing link between clinical observations and disease mechanisms.  In combination with clinical reasoning, this can  provide treating clinicians with decision support in several complementary ways.  

First, LHM can inform the clinicians about the values of the expert variables $\rvz^e(t)$ that cannot be observed in the clinical environment but are important for prognosis, choice of treatment, and anticipation of complications.  For example, understanding and predicting immune response is pivotal when deciding on immunosuppresive therapy in the treatment of COVID-19: an extreme immune response may lead to a
potentially fatal cytokine storm \cite{fajgenbaum2020cytokine}, but a suppressed immune response may be equally dangerous in case of (secondary) infection \cite{van2020corticosteroid,koehler2020defining}.  However, because immune response is not directly observable in the clinical environment, clinicians must rely on proxies such as C-reactive protein (CRP) for inflammation \cite{mortensen2001c}; by their very nature, such proxies are noisy and highly imperfect measures of the desired values.

Secondly,  LHM can provide the clinician with predictions of the disease progression given the treatments, enabling the  clinicians to design the best treatment plan for the patient at hand.   

Finally, LHM can bridge the gap between the laboratory and clinical  environments, helping to align model output with clinical reasoning, and thus to bring models to the patient bedside and also to foster translational research \cite{shillan2019use,fleuren2020machine}.

\section{Related works}
\label{sec:related_work}

\textbf{Hybrid models}. 
Hybrid models combine a given expert model with ML \cite{willard2020integrating}. Depending on the type and functionality of the expert model, various approaches have been proposed. 
\textit{Residual Models} and \textit{Ensembles} use expert models that can issue predictions directly \cite{liu2019multi,wang2017physics,wu2018physics, xu2015data,yao2018tensormol}. 
A Residual Model fits a ML model to the residuals of the expert model while an Ensemble  averages the ML and expert predictions. 
\textit{Feature Extraction} makes use of an expert model that  extracts useful features from the measurements \cite{karpatne2017physics}; an ML model then uses these features to make predictions. 
These  methods are not suitable for our setting because our expert model  is an ODE that governs the latent variables (\ref{eq:latent}); it does not issue predictions of measurements nor does it extract features that the ML model can use. Appendix \ref{app:related_work} Table \ref{tab:1} summarizes these approaches.

ML inspired by physics uses 
physical laws to guide the design of architectures \cite{schutt2017schnet,zepeda2019deep,thomas2018tensor}, loss functions \cite{yazdani2020systems,fioretto2020predicting}, and weight initialization \cite{read2019process}. Examples include  Hamiltonian neural networks \cite{greydanus2019hamiltonian,bertalan2019learning,zhong2019symplectic}, which reflect the conservation of energy. These models utilize  general physical laws rather than a specific expert model, and are rather different than the hybrid models discussed above.

\textbf{Neural ODEs}.
Neural ODEs approximate unknown ODEs by a neural network \cite{chen2018neural}, frequently using standard feed-forward networks.   ODE\textsuperscript{2}VAE uses an architecture with identity blocks to approximate second-order ODEs \cite{yildiz2019ode2vae} and GRU-ODE uses an architecture inspired by the Gated Recurrent Unit (GRU) \cite{de2019gru,cho2014learning}. 
Neural ODEs and extensions have achieved state-of-the-art performance in a variety of problems involving irregularly-sampled time series data \cite{rubanova2019latent,de2019gru, kidger2020neural}. We discuss other approaches to learning unknown ODEs from data in Appendix \ref{app:related_work}.

\textbf{Mechanistic models}. 
Mechanistic models are widely applied in sciences such as Physics \cite{stronge2018impact}, Epidemiology \cite{yang2020modified,gilchrist2002modeling}, and Pharmacology \cite{gesztelyi2012hill,katzung2012basic}. 
These models use ODEs to describe a system's continuous-time evolution under interventions. The intervention affects the system \textit{deterministically} through the governing ODEs; e.g., Equation (\ref{eq:expert}).
This deterministic notion of intervention effect is different from the probabilistic one adopted by the statistical causal inference literature \cite{rubin2005causal} (for a detailed discussion, see \cite{scholkopf2021toward} and Appendix \ref{app:related_work}).
LHM uses two ODEs to describe the disease progression under treatment: the known expert ODE $f^e$ and the data-driven Neural ODE $f^m$. 

\textbf{Latent variable models}. Latent variable models are widely used in  disease progression modeling \cite{wang2014unsupervised, alaaattentive}. These models attempt to infer a set of latent variables to predict  complex high-dimensional measurements. 
The latent variables sometimes have  high-level interpretations (e.g. cluster membership), but do not usually correspond to any well-defined and clinically meaningful physiological variable.  Moreover, without informative priors, the latent variables can usually be identified only up to certain transformations (e.g. permutation of cluster labels \cite{pakman2020neural}). By contrast, LHM involves medically meaningful expert variables driven by known governing equations and following informative priors.

\vspace{-.1in}

\section{Experiment and evaluation}

Here we present the results of two experimental studies, one with simulated data and one with real data.  In both experiments, we study the effect of dexamethasone treatment for COVID-19 patients.  Both studies are  modeled on the real-life treatment of COVID-19 patients in the ICU.

\vspace{-.1in}

\subsection{Simulation study} \label{sec:simulation} In this simulation, we use LHM to predict the results of a single dexamethasone treatment.  Each patient $i$ will receive a one-time treatment; with dosage $d_i \sim \text{uniform}[0,10]$ mg and time $s_i \sim \text{uniform}[0,14]$.  Our objective is to predict future measurements.

\textbf{Datasets}. We generated a variety of datasets to evaluate the model performance under different scenarios. 
To evaluate how the number of clinical measurements affects performance, we generated datasets with $D = 20, 40\ \mbox{or}\ 80$ observable physiological variables $\rvx$.  
For each dataset, we set the number of un-modeled states $\rvz^m$ according to the number of variables in $\rvx$ to be $M = D/10 = 2, 4\ \mbox{or}\ 8$ (respectively). (We made this choice to reflect the fact that a larger number of physiological variables often necessitates a larger number of un-modeled states.) 
We consider a time horizon of $T = 14$ days; this   is the median length of stay in hospital for Covid-19 patients \cite{rees2020covid}.  
After setting $M$ and $D$, we generate the data points within a dataset independently.

We use a pharmacological model adapted from \cite{dai2021prototype} that describes five expert variables ($E=5$) under dexamethasone treatment for COVID-19 patients. We use the same model in the real-data experiment. We specify the model and the expert variables in Appendix \ref{app:pharma}.
The un-modeled states $\rvz^m$ are governed by the nonlinear ODE $f^m$ shown in  equation (\ref{eq:latent});  the true physiological variables $\rvx$ are generated by the function $g$ in Equation (\ref{eq:physiological}).  The specifications of $f^m$ and $g$ are provided in Appendix \ref{app:simulation}. 
For each patient $i$, each of the components of its initial condition $\rvz_i(0)$ were independently drawn from an exponential distribution with rate $\lambda = 100$. 
Measurement noises are drawn independently  from $\epsilon_{it} \sim N(0, \sigma^2)$, for 
$\sigma=0.2, 0.4\ \mbox{or}\ 0.8$;  Equation (\ref{eq:error}). 
We first simulate  all daily measurements at $t = 1, 2, \ldots, T$, and then randomly remove measurements with probability $0.5$ to proxy the fact that measurements are made irregularly.  

\begin{figure}[t!]
  \centering
  \includegraphics[width=1\textwidth]{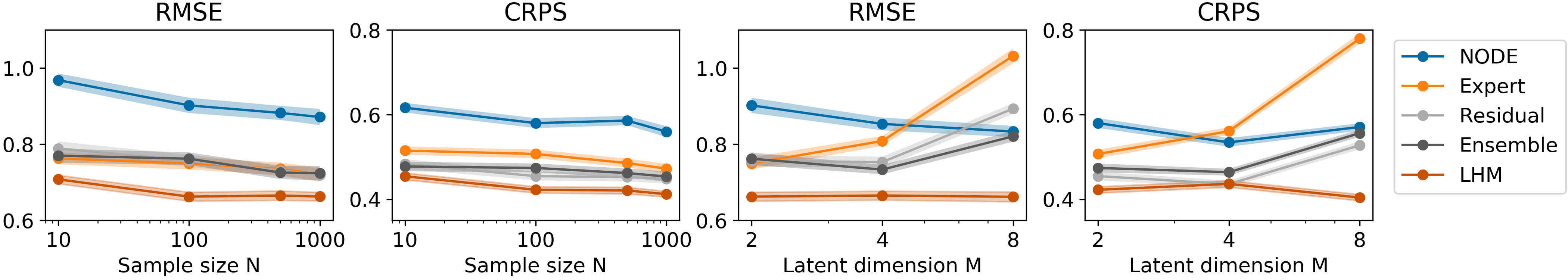}
  \caption{\footnotesize \textbf{Simulation Results}. Prediction accuracy on future measurements 
  $\mathcal{Y}[2:T]$ given the observed history $\mathcal{Y}[0:2]$ as measured by RMSE and CRPS.   The left 2 panels show the results under different training sample sizes $N_0$. The right 2 panels show the results for different numbers of un-modeled variables $M$.  The shaded areas represent 95\% confidence intervals.}
  \label{fig:exp}
\end{figure}

\textbf{Prediction task} For a given patient $i$, we use the measurements  $\mathcal{Y}[0:t_0]$ up to some time $t_0$ and the treatment plan for that particular patient to  {\em predict} the future measurements 
$\mathcal{Y}[t_0:T]$.  (Note that treatment may have occurred prior to time $t_0$ or may be planned following time $t_0$.)  To evaluate the performance under different lengths of observed history, we use $t_0 = 2, 5\ \mbox{or} \ 10$ days.   We evaluate the prediction accuracy according to Root Mean Squared Error (RMSE) and the uncertainty calibration using the Continuous Ranked Probability Score (CRPS).

\textbf{Training and Evaluation.}
We partition each dataset into a training set, a validation set, and a testing set. We consider training sets consisting of $N_0 = 10, 100, 500\ \mbox{or}\ 1000$ data points; each validation set has 100 data points and each testing set has 1000 data points.

\textbf{Benchmarks}.
We compare the performance of our method (LHM) with the performance of four other methods: latent Neural ODE (NODE), the original Pharmacology model (Expert), the residual model (Residual), and the ensemble model (Ensemble) of Expert and NODE, described below.
The details of the optimization and hyper-parameter settings are reported in Appendix \ref{app:simulation}.

{\bf NODE} involves $Z$ latent variables $\rvz(t) \in \mathbb{R}^Z$ whose evolutions are described by $\dot{\rvz}(t) = f^m(\rvz(t), \rva(t);\ \theta^m)$, where $f^m$ is a neural network with trainable weights $\theta^m$. 
The number of latent variables $Z$ is a \textit{hyper-parameter} that is set to be greater than $M+E$, which is the number of \textit{true} latent variables.\footnote{We found that the performance of NODE is not sensitive to the exact choice of $Z$ so long as it is sufficiently larger than $M+E$. (This is consistent with findings reported in the literature \cite{dupont2019augmented}.) } NODE predict the physiological variables as $\hat{\rvy}_N(t) = g_N(\rvz(t), \rva(t);\ \gamma_N)$, where $g_N$ is a a neural network with trainable weights $\gamma_N$.
{\bf Expert}  is given the true governing equation (\ref{eq:expert}), which describes the expert variables $\rvz^e(t)$, but no un-modeled latent variables. We use a neural network $g_E$ with trainable weights $\gamma_E$ to predict the physiological variables: $\hat{\rvy}_E(t) = g_E(\rvz^e(t), \rva(t);\ \gamma_E)$. 
{\bf Residual} Given a trained Expert model, we calculate its residuals $\rvr(t) = \rvy(t) - \hat{\rvy}_E(t)$. Then a NODE is trained to predict the residuals. The final prediction is $\hat{\rvy}_E(t) + \hat{\rvr}(t)$.
{\bf Ensemble }makes prediction as $w_{1t} \hat{\rvy}_N(t) + w_{2t} \hat{\rvy}_E(t)$, where $\hat{\rvy}_N(t)$ and $\hat{\rvy}_E(t)$ are the predictions issued by NODE and Expert respectively. The ensemble weights $w_{1t}$ and $w_{2t}$ are learned on the validation set to minimize the prediction error.


\textbf{Results.} Figure \ref{fig:exp} shows the predictive performance with $t_0=2$ and $\sigma=0.2$ (results for other settings are reported in Appendix \ref{app:simulation}). LHM achieves the best overall performance.  Expert does not perform well because it leaves out $\rvz^m(t)$. As expected, its performance significantly degrades as the number $M$ of latent variables increases. NODE is more robust to  increases in $M$ because it treats the number of latent variables as a hyper-parameter. Nevertheless, NODE is less sample efficient and it achieves worse performance for sample sizes up to 1000 (left half). Both Residual and Ensemble achieve performance gains over NODE and Expert alone, but they under-perform LHM because they perform averaging directly in the output space rather than trying to infer $\rvz^m$ in the latent space.

\subsection{Real-data experiments}
\label{sec:real_data}

\begin{table}[t]
  \caption{Prediction accuracy (RMSE) on COVID-19 intensive care data under different training sample sizes $N$. Prediction horizon $H=24$ hours. The standard deviations are shown in the brackets.}
  \label{real-table}
  \footnotesize
  \centering
\begin{tabular}{@{}lcccc@{}}
\toprule
Method \textbackslash $N_0$ & 100          & 250          & 500          & 1000         \\ \midrule
Expert                   & 0.718 (0.71) & 0.704 (0.02) & 0.702 (0.02) & 0.713 (0.01) \\
Residual                 & 0.958 (0.63) & 1.003 (0.03) & 0.717 (0.05) & 0.635 (0.04) \\
Ensemble                 & 0.707 (0.60) & 0.657 (0.05) & 0.628 (0.05) & 0.599 (0.05) \\
NODE                     & 0.662 (0.65) & 0.659 (0.02) & 0.644 (0.05) & 0.650 (0.04) \\
ODE2VAE                  & 0.674 (0.62) & 0.666 (0.02) & 0.643 (0.02) & 0.619 (0.02) \\
GRU-ODE                  & 0.722 (0.60) & 0.673 (0.05) & 0.623 (0.05) & 0.601 (0.05) \\
Time LSTM                & 0.706 (0.63) & 0.649 (0.03) & 0.600 (0.03) & 0.631 (0.02) \\
LHM                      & \textbf{0.633 (0.51)} & \textbf{0.605 (0.02)} & \textbf{0.529 (0.02)} & \textbf{0.511 (0.02)} \\ \bottomrule
\end{tabular}
\end{table}

In this experiment, we use real data to evaluate the predictive performance of LHM and to  illustrate its utility for decision support in a realistic clinical setting that closely tracks the actual treatment of COVID-19 patients in ICU.

\textbf{Dataset}. We used data from the Dutch Data Warehouse (DDW), a multicenter and full-admission anonymized electronic health records database of critically ill COVID-19 patients \cite{fleuren2021large}. 
Up until March 2021, DDW has collected the health trajectories for 3464 patients admitted to intensive care units (ICU) across the Netherlands. 
However, even if we use the entire DDW for training, the sample size is still relatively small compared to what is typically used by ML (tens or  hundreds of thousands of samples \cite{johnson2016mimic}). 
Furthermore, patients are even scarcer at the early stage of pandemic, arguably when a decision support tool is most needed: only 607 patients were admitted at the first peak (by April 2020).
After applying the eligibility criterion detailed in Appendix \ref{app:real_data}, we obtained a dataset of \textbf{2097} patients whose disease progression is characterized by an irregularly-sampled time series of \textbf{27} physiological variables ({Appendix} \ref{app:real_data}). 
These variables capture the vital signals, respiratory mechanics, and biomarkers that are crucial for clinical decisions. In addition, we also included \textbf{11} static variables that are known to affect the progression of COVID-19, e.g. BMI ({Appendix} \ref{app:real_data}).

\textbf{Prediction task}. Denote $t_0$ as the time when the patient received the first dose of dexamethasone (we set $t_0=24$ for untreated patients). We use the history up to 24 hours before $t_0$,  $\mathcal{Y}[t_0-24:t_0]$, to predict the future $\mathcal{Y}[t_0:t_0+24H]$ over a time horizon $H = 1, 3\ \mbox{or}\ 7$ days. We use $N_0 = 100, 250, 500\ \mbox{or}\ 1000$ patients for training, {97} for validation, and {1000} for testing. The pharmacological model and the prior distribution of  expert variables are detailed in Appendix \ref{app:pharma}.

\textbf{Benchmarks}. In addition to all the benchmarks introduced in Section \ref{sec:simulation}, we compared the results with two extensions of NODE, GRU-ODE and ODE\textsuperscript{2}VAE, which achieved strong performance in medical time series prediction  \cite{yildiz2019ode2vae, de2019gru}. We also used the Time LSTM as a strong baseline \cite{baytas2017patient}. 

\textbf{Results}. The main results are shown in Table \ref{real-table} (additional results are shown in Appendix \ref{app:real_data}). LHM consistently outperformed the benchmarks. 
Its performance with $N_0=100$ samples is close to the pure ML approaches' performance with $N_0=500$ samples.
As the sample size increases from $100$ to $1000$, the predictive accuracy of LHM improves by $19\%$ while Time LSTM improves by $11\%$ and NODE by less than $5\%$.
A larger improvement rate suggests LHM adapts to the newly available data faster, which is important when the samples are scarce. 
As expected, the standalone expert model achieved poor performance because it is unable to capture the full array of clinical measurements.

\begin{figure}[t!]
  \centering
  \includegraphics[width=0.99\textwidth,trim={0.48cm 0 0 0},clip]{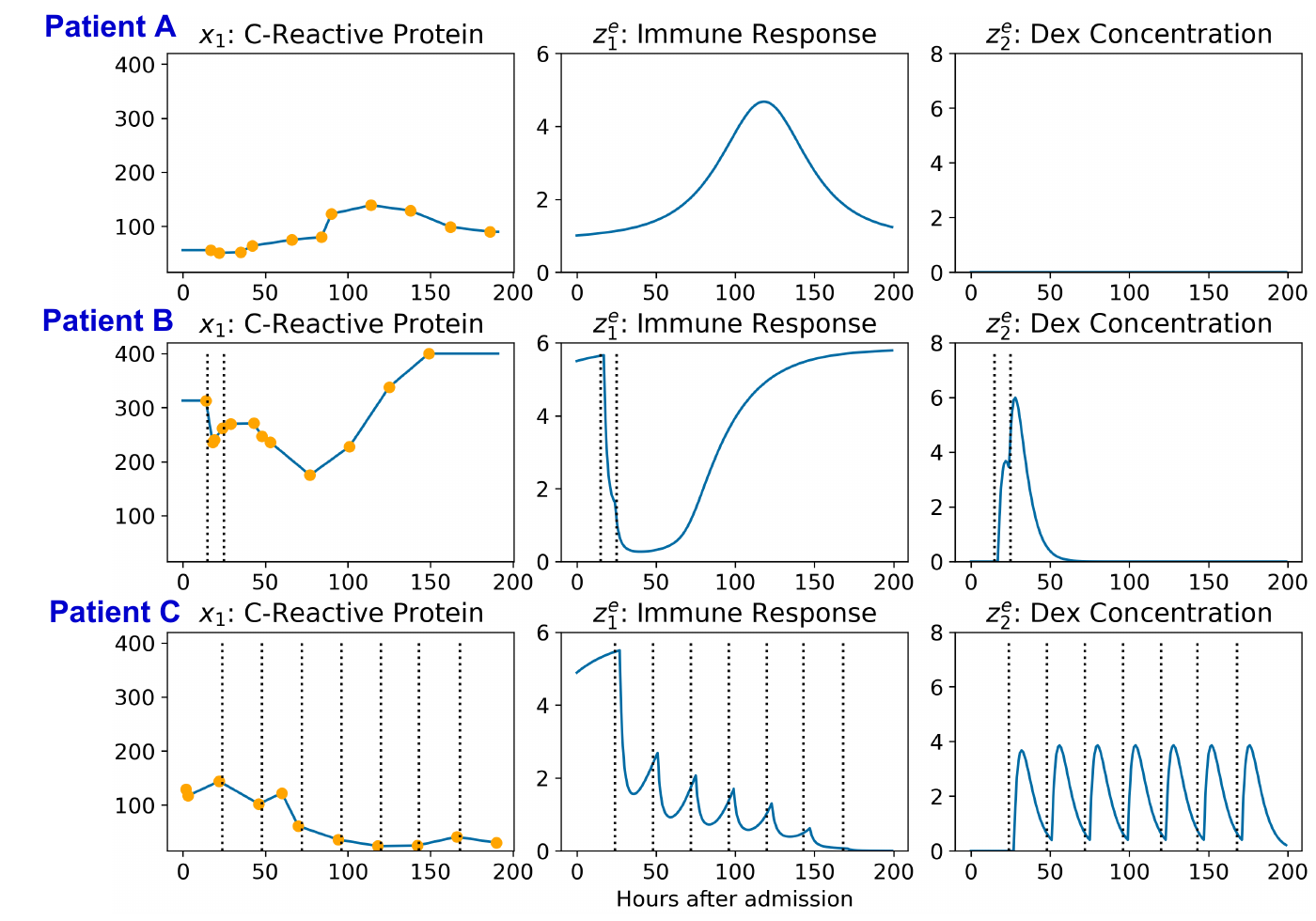}
  \caption{\footnotesize \textbf{The observed measurements and the inferred expert variables for three illustrative patients}. Left: The observed physiological variable $x_1(t)$: C-Reactive Protein. Middle: The inferred expert variable $z^e_1(t)$: the immune response to viral infection. Right: The inferred expert variable $z^e_2(t)$: dexamethasone concentration. Vertical dotted lines mark the times of dexamethasone injections.}
  \label{fig:real_interp}
\end{figure}

\textbf{LHM in action}. Here we show how LHM can support clinical decisions beyond predicting future clinical measurements. 
Managing the level of immune response is pivotal when deciding on immunosuppresive therapy for COVID-19 patients \cite{fajgenbaum2020cytokine,van2020corticosteroid,koehler2020defining}.
This is a challenging task because the ``right'' level of immune response varies across patients \cite{keller2020effect}: for most patients, we would like to reduce the immune response to avoid cytokine storm and consequent organ failure \cite{fajgenbaum2020cytokine}, but for patients with other infections (e.g., a secondary bacterial or fungal infection), we would like to keep their immune systems activated \cite{koehler2020defining,cox2020co}. 
Because immune response is not directly observable in the clinical settings, clinicians resort to unspecific inflammatory markers such as C-Reactive Protein (CRP) \cite{mortensen2001c}. However, better markers of the immune response such as the cytokine Type I IFNs can be measured in the laboratory setting and have been included in the  pharmacological model as an expert variable $z^e_1$.
Moreover, the immune response is affected by dexamethasone concentrations in the lung tissue (the expert variable $z^e_2$). These concentrations are not easily or routinely measured in a clinical setting, and are therefore not available to clinicians.

Figure \ref{fig:real_interp} shows the \textit{measurements} of CRP $\rvx_1$, and the \textit{inferred} immune response $z^e_1$ and dexamethasone concentration $z^e_2$. The two expert variables are inferred by a trained LHM using the first five days of observations. We selected three representative patients based on the treatment regimen they received. 
Patient A is representative of the 59.8\% of the population who did not receive dexamethasone; Patient C is representative of the 12.2\% who received dexamethasone according to the guidelines \cite{guideline}, and Patient B is representative of the 28.0\% of the population who received dexamethasone but whose treatment was not according to the guidelines.

For patient A, the initial level of CRP was moderately high, but then it rose and peaked at about 100 hours after admission. In the absence of contraindications, a clinician might begin dexamethasone treatment at this point, but LHM predicts that immune activity $z^e_1$ would decrease afterwards even without treatment. (The right panel is blank because  dexamethasone was never administered.)

Patient B was admitted to the ICU with a very high level of the inflammatory marker CRP. Two doses of dexamethasone were given in rapid succession, preceding a decline in both CRP $x_1$ and immune activity $z^e_1$. However, after the dexamethasone depleted in the patient's body, the expert model predicts that the immune response will pick up again. This is reflected in the re-occurrence of the high CRP level. 

Patient C was admitted to the ICU with a moderately high level of CRP. LHM also inferred a high level of immune activity $z^e_1$ at the initial stage. Inflammation was greatly reduced after dexamethasone treatments, which has an immunosuppressive effect; so $x_1$ and $z^e_1$ display the same downward trend. Because dexamethasone concentrations falls rapidly within 24 hours after treatment, it was repeated at 24 hour intervals, as seen in the right panel. This pattern is clinically expected under the treatment regimen.

\section{Discussion and future work}
\label{sec:discussion}
This paper has focused  on  a single disease (COVID-19),  a single treatment (dexamethasone), and a single expert model.  The ultimate goal is to build a model that encompasses a variety of diseases, a variety of treatments and multiple expert models.  This is a challenge for future work.

\begin{ack}

This work was supported by the Office of Naval Research (ONR) and the NSF (Grant number:1722516).

We would like to thank the following individuals for providing domain expertise in Pharmacology and designing the expert ODEs. 

Bernhard Steiert, PhD, Pharmaceutical Sciences, Roche Pharma Research and Early Development (pRED), Roche Innovation Center Basel, Basel, Switzerland

Richard Peck, MA, MB, BChir, FFPM, FRCP, Pharmaceutical Sciences, Roche Pharma Research and Early Development (pRED), Roche Innovation Center Basel, Basel, Switzerland

We would like to thank the following individuals for curating and sharing the Dutch Data Warehouse (DDW) for critically ill COVID-19 patients.

\textit{From collaborating hospitals having shared data:}

Diederik Gommers, MD, PhD, Department of Intensive Care, Erasmus Medical Center, Rotterdam, The Netherlands, 

Olaf L. Cremer, MD, PhD, Intensive Care, UMC Utrecht, Utrecht, The Netherlands, 

Rob J. Bosman, MD, ICU, OLVG, Amsterdam, The Netherlands, 

Sander Rigter, MD, Department of Anesthesiology and Intensive Care, St. Antonius Hospital, Nieuwegein, The Netherlands, 

Evert-Jan Wils, MD, PhD, Department of Intensive Care, Franciscus Gasthuis \& Vlietland, Rotterdam, The Netherlands, 

Tim Frenzel, MD, PhD, Department of Intensive Care Medicine, Radboud University Medical Center, Nijmegen, The Netherlands, 

Dave A. Dongelmans, MD, PhD, Department of Intensive Care Medicine, Amsterdam UMC, Amsterdam, The Netherlands, 

Remko de Jong, MD, Intensive Care, Bovenij Ziekenhuis, Amsterdam, The Netherlands, 

Marco Peters, MD, Intensive Care, Canisius Wilhelmina Ziekenhuis, Nijmegen, The Netherlands, 

Marlijn J.A Kamps, MD, Intensive Care, Catharina Ziekenhuis Eindhoven, Eindhoven, The Netherlands, 

Dharmanand Ramnarain, MD, Department of Intensive Care, ETZ Tilburg, Tilburg, The Netherlands, 

Ralph Nowitzky, MD, Intensive Care, HagaZiekenhuis, Den Haag, The Netherlands, 

Fleur G.C.A. Nooteboom, MD, Intensive Care, Laurentius Ziekenhuis, Roermond, The Netherlands, 

Wouter de Ruijter, MD, PhD, Department of Intensive Care Medicine, Northwest Clinics, Alkmaar, The Netherlands, 

Louise C. Urlings-Strop, MD, PhD, Intensive Care, Reinier de Graaf Gasthuis, Delft, The Netherlands, 

Ellen G.M. Smit, MD, Intensive Care, Spaarne Gasthuis, Haarlem en Hoofddorp, The Netherlands, 

D. Jannet Mehagnoul-Schipper, MD, PhD, Intensive Care, VieCuri Medisch Centrum, Venlo, The Netherlands, 

Julia Koeter, MD, Intensive Care, Canisius Wilhelmina Ziekenhuis, Nijmegen, The Netherlands, 

Tom Dormans, MD, PhD, Intensive care, Zuyderland MC, Heerlen, The Netherlands, 

Cornelis P.C. de Jager, MD, PhD, Department of Intensive Care, Jeroen Bosch Ziekenhuis, Den Bosch, The Netherlands, 

Stefaan H.A. Hendriks, MD, Intensive Care, Albert Schweitzerziekenhuis, Dordrecht, The Netherlands, 

Sefanja Achterberg, MD, PhD, ICU, Haaglanden Medisch Centrum, Den Haag, The Netherlands, 

Evelien Oostdijk, MD, PhD, ICU, Maasstad Ziekenhuis Rotterdam, Rotterdam, The Netherlands, 

Auke C. Reidinga, MD, ICU, SEH, BWC, Martiniziekenhuis, Groningen, The Netherlands, 

Barbara Festen-Spanjer, MD, Intensive Care, Ziekenhuis Gelderse Vallei, Ede, The Netherlands, 

Gert B. Brunnekreef, MD, Department of Intensive Care, Ziekenhuisgroep Twente, Almelo, The Netherlands, 

Alexander D. Cornet, MD, PhD, FRCP, Department of Intensive Care, Medisch Spectrum Twente, Enschede, The Netherlands, 

Walter van den Tempel, MD, Department of Intensive Care, Ikazia Ziekenhuis Rotterdam, Rotterdam, The Netherlands, 

Age D. Boelens, MD, Anesthesiology, Antonius Ziekenhuis Sneek, Sneek, The Netherlands, 

Peter Koetsier, MD, Intensive Care, Medisch Centrum Leeuwarden, Leeuwarden, The Netherlands, 

Judith Lens, MD, ICU, IJsselland Ziekenhuis, Capelle aan den IJssel, The Netherlands, 

Roger van Rietschote, Business Intelligence, Haaglanden MC,  Den Haag,The Netherlands, 

Harald J. Faber, MD, ICU, WZA, Assen, The Netherlands, 

A. Karakus, MD, Department of Intensive Care, Diakonessenhuis Hospital, Utrecht, The Netherlands, 

Robert Entjes, MD, Department of Intensive Care, Admiraal De Ruyter Ziekenhuis, Goes, The Netherlands, 

Paul de Jong, MD, Department of Anesthesia and Intensive Care, Slingeland Ziekenhuis, Doetinchem, The Netherlands, 

Thijs C.D. Rettig, MD, PhD, Department of Intensive Care, Amphia Ziekenhuis, Breda, The Netherlands, 

M.C. Reuland, MD, Department of Intensive Care Medicine, Amsterdam UMC, Universiteit van Amsterdam, Amsterdam, The Netherlands, 

Laura van Manen, MD, Department of Intensive Care, BovenIJ Ziekenhuis, Amsterdam, The Netherlands, 

Leon Montenij, MD, PhD, Department of Anesthesiology, Pain Management and Intensive Care, Catharina Ziekenhuis Eindhoven, Eindhoven, The Netherlands, 

Jasper van Bommel, MD, PhD, Department of Intensive Care, Erasmus Medical Center, Rotterdam, The Netherlands, 

Roy van den Berg, Department of Intensive Care, ETZ Tilburg, Tilburg, The Netherlands, 

Ellen van Geest, Department of ICMT, Haga Ziekenhuis, Den Haag, The Netherlands, 

Anisa Hana, MD, PhD, Intensive Care, Laurentius Ziekenhuis, Roermond, The Netherlands, 

B. van den Bogaard, MD, PhD, ICU, OLVG, Amsterdam, The Netherlands, 

Prof. Peter Pickkers, Department of Intensive Care Medicine, Radboud University Medical Centre, Nijmegen, The Netherlands, 

Pim van der Heiden, MD, PhD, Intensive Care, Reinier de Graaf Gasthuis, Delft, The Netherlands, 

Claudia (C.W.) van Gemeren, MD, Intensive Care, Spaarne Gasthuis, Haarlem en Hoofddorp, The Netherlands, 

Arend Jan Meinders, MD, Department of Internal Medicine and Intensive Care, St Antonius Hospital, Nieuwegein, The Netherlands, 

Martha de Bruin, MD, Department of Intensive Care, Franciscus Gasthuis \& Vlietland, Rotterdam, The Netherlands, 

Emma Rademaker, MD, MSc, Department of Intensive Care, UMC Utrecht, Utrecht, The Netherlands, 

Frits H.M. van Osch, PhD, Department of Clinical Epidemiology, VieCuri Medisch Centrum, Venlo, The Netherlands, 

Martijn de Kruif, MD, PhD, Department of Pulmonology, Zuyderland MC, Heerlen, The Netherlands, 

Nicolas Schroten, MD, Intensive Care, Albert Schweitzerziekenhuis, Dordrecht, The Netherlands, 

Klaas Sierk Arnold, MD, Anesthesiology, Antonius Ziekenhuis Sneek, Sneek, The Netherlands, 

J.W. Fijen, MD, PhD, Department of Intensive Care, Diakonessenhuis Hospital, Utrecht, The Netherland, 

Jacomar J.M. van Koesveld, MD, ICU, IJsselland Ziekenhuis, Capelle aan den IJssel, The Netherlands, 

Koen S. Simons, MD, PhD, Department of Intensive Care, Jeroen Bosch Ziekenhuis, Den Bosch, The Netherlands, 

Joost Labout, MD, PhD, ICU, Maasstad Ziekenhuis Rotterdam, The Netherlands, 

Bart van de Gaauw, MD, Martiniziekenhuis, Groningen, The Netherlands, 

Michael Kuiper, Intensive Care, Medisch Centrum Leeuwarden, Leeuwarden, The Netherlands, 

Albertus Beishuizen, MD, PhD, Department of Intensive Care, Medisch Spectrum Twente, Enschede, The Netherlands, 

Dennis Geutjes, Department of Information Technology, Slingeland Ziekenhuis, Doetinchem, The Netherlands, 

Johan Lutisan, MD, ICU, WZA, Assen, The Netherlands, 

Bart P. Grady, MD, PhD, Department of Intensive Care, Ziekenhuisgroep Twente, Almelo, The Netherlands, 

Remko van den Akker, Intensive Care, Adrz, Goes, The Netherlands, 

Sesmu Arbous, MD, PhD, Intensivist, LUMC, Leiden, The Netherlands, 

Tom A. Rijpstra, MD, Department of Intensive Care, Amphia Ziekenhuis, Breda, The Netherlands, 

Roos Renckens, MD, PhD, Department of Internal Medicine, Northwest Clinics, Alkmaar, the Netherlands, 

\textit{From collaborating hospitals having signed the data sharing agreement:}

Daniël Pretorius, MD, Department of Intensive Care Medicine, Hospital St Jansdal, Harderwijk, The Netherlands, 

Menno Beukema, MD, Department of Intensive Care, Streekziekenhuis Koningin Beatrix, Winterswijk, The Netherlands, 

Bram Simons, MD, Intensive Care, Bravis Ziekenhuis, Bergen op Zoom en Roosendaal, The Netherlands, 

A.A. Rijkeboer, MD, ICU, Flevoziekenhuis, Almere, The Netherlands, 

Marcel Aries, MD, PhD, MUMC+, University Maastricht, Maastricht, The Netherlands, 

Niels C. Gritters van den Oever, MD, Intensive Care, Treant Zorggroep, Emmen, The Netherlands, 

Martijn van Tellingen, MD, EDIC, Department of Intensive Care Medicine, afdeling Intensive Care, ziekenhuis Tjongerschans, Heerenveen, The Netherlands, 

Annemieke Dijkstra, MD, Department of Intensive Care Medicine, Het Van Weel-Bethesda Ziekenhuis, Dirksland, The Netherlands, 

Rutger van Raalte, Department of Intensive Care, Tergooi hospital, Hilversum, The Netherlands, 

\textit{From the Laboratory for Critical Care Computational Intelligence:
}

Tariq A. Dam, MD, Department of Intensive Care Medicine, Laboratory for Critical Care Computational Intelligence, Amsterdam Medical Data Science, Amsterdam UMC, Vrije Universiteit, Amsterdam, The Netherlands, 

Martin E. Haan, MD, Department of Intensive Care Medicine, Laboratory for Critical Care Computational Intelligence, Amsterdam Medical Data Science, Amsterdam UMC, Vrije Universiteit, Amsterdam, The Netherlands, 

Mark Hoogendoorn, PhD, Quantitative Data Analytics Group, Department of Computer Science, Faculty of Science, VU University, Amsterdam, The Netherlands, 

Armand R.J. Girbes, MD, PhD, EDIC, Department of Intensive Care Medicine, Laboratory for Critical Care Computational Intelligence, Amsterdam Medical Data Science, Amsterdam UMC, Vrije Universiteit, Amsterdam, The Netherlands, 

Paul W.G. Elbers, MD, PhD, EDIC, Department of Intensive Care Medicine, Laboratory for Critical Care Computational Intelligence, Amsterdam Medical Data Science, Amsterdam UMC, Vrije Universiteit, Amsterdam, The Netherlands, 

Patrick J. Thoral, MD, EDIC, Department of Intensive Care Medicine, Laboratory for Critical Care Computational Intelligence, Amsterdam Medical Data Science, Amsterdam UMC, Vrije Universiteit, Amsterdam, The Netherlands, 

Dagmar M. Ouweneel, PhD, Department of Intensive Care Medicine, Laboratory for Critical Care Computational Intelligence, Amsterdam Medical Data Science, Amsterdam UMC, Vrije Universiteit, Amsterdam, The Netherlands, 

Ronald Driessen, Department of Intensive Care Medicine, Laboratory for Critical Care Computational Intelligence, Amsterdam Medical Data Science, Amsterdam UMC, Vrije Universiteit, Amsterdam, The Netherlands, 

Jan Peppink, Department of Intensive Care Medicine, Laboratory for Critical Care Computational Intelligence, Amsterdam Medical Data Science, Amsterdam UMC, Vrije Universiteit, Amsterdam, The Netherlands, 

H.J. de Grooth, MD, PhD, Department of Intensive Care Medicine, Laboratory for Critical Care Computational Intelligence, Amsterdam Medical Data Science, Amsterdam UMC, Vrije Universiteit, Amsterdam, The Netherlands, 

\textit{From Pacmed:}

Robbert C.A. Lalisang, MD, Pacmed, Amsterdam, The Netherlands, 

Michele Tonutti, MRes, Pacmed, Amsterdam, The Netherlands, 

Daan P. de Bruin, MSc, Pacmed, Amsterdam, The Netherlands, 

Sebastiaan J.J. Vonk, MSc, Pacmed, Amsterdam, The Netherlands, 

Mattia Fornasa, PhD, Pacmed, Amsterdam, The Netherlands, 

Tomas Machado, Pacmed, Amsterdam, The Netherlands, 

Michael de Neree tot Babberich, Pacmed, Amsterdam, The Netherlands, 

Olivier Thijssens, MSc, Pacmed, Amsterdam, The Netherlands, 

Lot Wagemakers, Pacmed, Amsterdam, The Netherlands, 

Hilde G.A. van der Pol, Pacmed, Amsterdam, The Netherlands, 

Tom Hendriks, Pacmed, Amsterdam, The Netherlands, 

Julie Berend, Pacmed, Amsterdam, The Netherlands, 

Virginia Ceni Silva, Pacmed, Amsterdam, The Netherlands, 

Robert F.J. Kullberg, MD, Pacmed, Amsterdam, The Netherlands, 

Taco Houwert, MSc, Pacmed, Amsterdam, The Netherlands, 

Hidde Hovenkamp, MSc, Pacmed, Amsterdam, The Netherlands, 

Roberto Noorduijn Londono, MSc, Pacmed, Amsterdam, The Netherlands, 

Davide Quintarelli, MSc, Pacmed, Amsterdam, The Netherlands, 

Martijn G. Scholtemeijer, MD, Pacmed, Amsterdam, The Netherlands, 

Aletta A. de Beer, MSc, Pacmed, Amsterdam, The Netherlands, 

Giovanni Cina, PhD, Pacmed, Amsterdam, The Netherlands, 

Willem E. Herter, BSc, Pacmed, Amsterdam, The Netherlands, 

Adam Izdebski, Pacmed, Amsterdam, The Netherlands, 

\textit{From RCCnet:}

Leo Heunks, MD, PhD, Department of Intensive Care Medicine, Amsterdam Medical Data Science, Amsterdam UMC, Vrije Universiteit, Amsterdam, The Netherlands, 

Nicole Juffermans, MD, PhD, ICU, OLVG, Amsterdam, The Netherlands, 

Arjen J.C. Slooter, MD, PhD, Department of Intensive Care Medicine, UMC Utrecht, Utrecht University, Utrecht, the Netherlands, 

\textit{From other collaborating partners:}

Martijn Beudel, MD, PhD, Department of Neurology, Amsterdam UMC, Universiteit van Amsterdam, Amsterdam, The Netherlands,

\end{ack}

\clearpage

\appendix

\section{Appendix}

\textbf{Summary of Appendices.}

Each section can be read independently.

\begin{itemize}
    \item \ref{app:pharma}: The pharmacological model for dexamethasone.
    \item \ref{app:impact}: Medical importance and impact.
    \item \ref{app:optimization}: Optimization and gradient calculation.
    \item \ref{app:simulation}: Details about the simulation study.
    \item \ref{app:related_work}: Extended related works.
    \item \ref{app:real_data}: Details about the real data experiment.
    \item \ref{app:extensions}: Practical extensions of LHM.
\end{itemize}

\subsection{The pharmacological model for dexamethasone}
\label{app:pharma}

\begin{figure}[t!]
  \centering
  \includegraphics[width=0.80\textwidth]{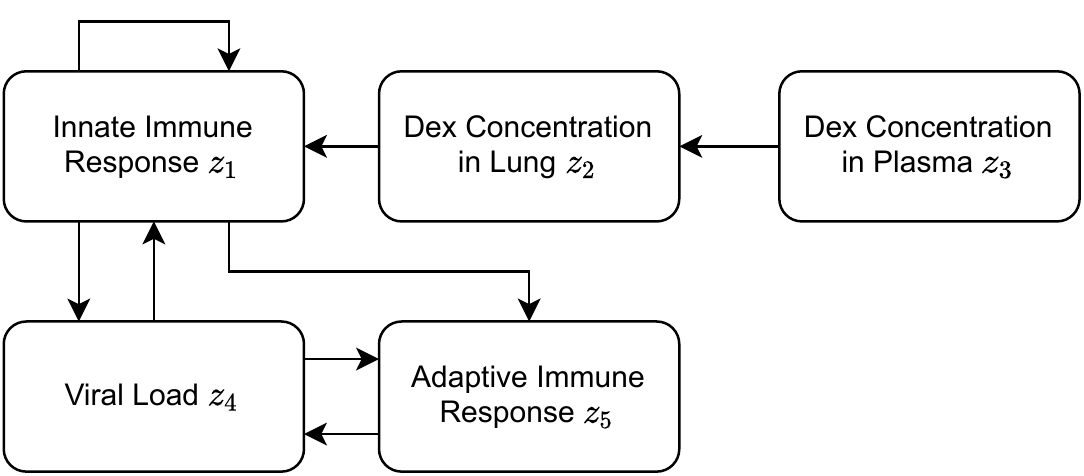}
  \caption{\footnotesize The expert variables and their temporal interactions as described by the expert ODE.}
  \label{fig:Pharma_diagram}
\end{figure}

The expert ODE we used is adapted from \cite{dai2021prototype}. As illustrated in Figure \ref{fig:Pharma_diagram}, it involves five expert variables $z_1$ to $z_5$ (the superscript $e$ is omitted for brevity). The $z_1$ represents the innate immune response to viral infection (measured in the laboratory using Type I IFNs \cite{khader2020targeting,abbas2014cellular}). 
The  $z_2$ and  $z_3$ represent the concentration of dexamethasone in lung tissue and plasma respectively. 
The $z_4$ represents the viral load and $z_5$ represents the adaptive immune response (measured in the laboratory using Cytotoxic T cells \cite{abbas2014cellular}). 

The expert model that describes these variables are developed based on specialized knowledge and laboratory experiments. Firstly, the immune responses and viral replication are modeled as:
\begin{align}
\label{eq:expert_model}
    \dot{z_1} &= k_{IR} \cdot z_4 + k_{PF}\cdot z_4\cdot z_1 - k_{O}\cdot z_1 + \frac{E_{max}\cdot z_1^{h_P}}{EC_{50}^{h_P} +  z_1^{h_P}} - k_{Dex}\cdot z_1\cdot z_2 \\
    \dot{z_4} &= k_{DP}\cdot z_4 - k_{IIR}\cdot z_4\cdot z_1 - k_{DC}\cdot z_4\cdot z_5^{h_C}\\
    \dot{z_5} &= k_{1}\cdot z_1
\end{align}
The five terms in the first Equation for $z_1$ captures the initial immune reaction to the virus, the physiological positive feedback, the immune cell mortality, the pathological positive feedback, and the effect of dexamethasone \cite{dai2021prototype}. The second equation of $z_4$ captures the viral replication, and the effect of innate and adaptive immune systems on the virus. The last equation captures the adaptive immune response triggered by the innate immune response \cite{abbas2014cellular}. The unknown coefficients $k_{IR}$, $k_{PF}$, $k_{O}$, $E_{max}$, $h_P$, $k_{Dex}$, $k_{DP}$, $k_{IIR}$, $k_{DC}$, $h_C$ are positive real numbers. 

The concentration of dexamethasone ($z_2$, $z_3$) is described by a standard two-compartmental pharmacokinetics model  \cite{metzler1971usefulness, kramer1974pharmacokinetics}:
\begin{align}
\label{eq:expert_model2}
    \dot{z_2} &= - k_{2}\cdot z_2 + k_{3}\cdot z_3 \\
    \dot{z_3} &= - k_{3}\cdot z_3
\end{align}
The coefficients $k_{2}$, $k_{3}$ are positive real numbers. 
In the literature, it is often assumed for simplicity that the treatment is given at time $t=0$, and the initial condition of the plasma concentration $z_3(0)$ corresponds to the dosage \cite{earp2008pharmacokinetics}. Since the plasma concentration $z_3$ decays exponentially over time, we can equivalently express it as a sum of exponentials: $z_3(t) = \sum_{i} d_i \cdot I(t > t_i) \cdot \text{exp}(k_{3} (t_i - t)) $ when dosages $d_i$ are given at time $t_i$, $i\ge1$. The function $I(\cdot)$ is an indicator function.

\textbf{Prior distribution for real-data experiment}. The initial condition $\rvz(0)$ corresponds to the patient state at the time of ICU admission. Since dexamethasone is generally administered \textit{during} the ICU stay \cite{guideline}, its concentration at admission  should be very close to zero. Hence we use an exponential distribution with rate $\lambda=100$ as the prior of $z_2(0)$ and $z_3(0)$. On the other hand, the immune response and viral load may vary across patients greatly. To allows for more heterogeneity, we use an exponential distribution with rate $\lambda=0.1$ as the prior of $z_1(0)$, $z_4(0)$, and $z_5(0)$. The exponential distribution also reflects the positivity of the expert variables because it has a positive support.

\subsection{Medical importance and impact}
\label{app:impact}

\subsubsection{Medical importance of LHM}

Integration of machine learning (ML) and pathophysiology is a major challenge in adopting ML models in a clinical setting. While clinicians seek to understand the mechanisms that drive disease progression for prognosis and treatment allocation, machine learning models do not currently provide such disease dynamics. With these dynamics, however, model results would translate to clinically interpretable concepts, would resonate with clinicians, and could then support clinical decision making. 

In addition, disease dynamics are indispensable for the clinical interpretation of predictive modeling results. Predictive modeling has taken flight in the medical field, but many models are left stranded because clinicians can solely rely on feature importance and no mechanistic interpretation of the results. Such an interpretation, however, will increase clinicians’ trust in these models and expedite their use in clinical practice. 

Lastly, the relationship between fundamental and clinical research may yield novel hypotheses and foster subsequent research. There is a gap between benchwork and the bedside. Bridging this gap with ML and interpretable models could reveal novel relationships and could inspire research both ways. Overall, ML with a mechanistic interpretation can provide the next big step in medical data science and can help bring these models to the bedside.

\subsubsection{Dexamethasone in COVID-19}

Coronavirus disease 2019 (COVID-19) was an unknown disease to intensive care clinicians worldwide. Both the natural course of the disease as well as optimal treatment were unknown throughout the onset of the pandemic. Since inflammatory organ injury appeared to play an important role in the pathophysiology of COVID-19, glucocorticoids were proposed to mitigate the damaging effects of the immune system \cite{moore2020cytokine}. In particular, Dexamethasone treatment has been shown to reduce mortality in patients on invasive mechanical ventilation or oxygen alone in the RECOVERY trial \cite{recovery2021dexamethasone}. Moreover, the CoDEX trial demonstrated an increase in the number of ventilator free days with Dexamethasone treatment in moderate to severe COVID-19 acute respiratory distress syndrome (ARDS)\cite{tomazini2020effect}. As a result, COVID-19 treatment guidelines recommend Dexamethasone treatment in these settings \cite{guideline}.

Although beneficial effects have been shown of Dexamethasone on a group level, individual response to treatment remains unknown. Knowing this response would help clinicians to anticipate complications, to improve individualized prognosis, and potentially determine beneficial treatments in these patients. Moreover, clinicians could identify patients in which Dexamethasone has a desired effect and in which patient it may not. For example, in the case of coinfection, Dexamethasone may be discontinued in selected patients. Lastly, these models can identify novel mechanistic pathways in COVID-19 patients that can inspire both fundamental and clinical research. Taken together, individualized disease progression in response to Dexamethasone treatment would bring about a large step forward in COVID-19 research.

\subsubsection{Potential negative impact}
Any decision support system could be used negatively if the user intentionally chooses to worsen the outcome. This is very unlikely in our case because the intended users of LHM are clinicians.

\subsection{Optimization and gradient calculation}
\label{app:optimization}

We optimize ELBO by stochastic gradient ascent using the ADAM optimizer \cite{kingma2014adam}. The gradient calculation is enabled by the following two methods.

\textbf{Reparameterization}. To evaluate the ELBO, we need to take samples from the variational distribution $\mathbb{Q}_\phi$. Here we use the Gaussian reparameterization in all sampling steps to obtain the gradients for the encoder \cite{kingma2013auto}.

\textbf{Gradient for ODE}. We use the torchdiffeq library to calculate the gradient with respect to the ODE solutions \cite{chen2018neural}. A variety of ODE solvers are available in the library, we used the adams solver, which is an adaptive step size solver.

\subsection{Simulation study}
\label{app:simulation}

\subsubsection{Data generation details}

We generated a variety of datasets to evaluate the model performance under different scenarios. 
To evaluate how the number of clinical measurements affects performance, we generated datasets with $D = 20, 40\ \mbox{or}\ 80$ measurable physiological variables $\rvx$.  
For the pharmacological model, we used the model provided in Appendix \ref{app:pharma}, which involves five inter-related variables ($E=5$). We set the coefficients $h_P = h_C = 2$ and the rest to be one. 

For each dataset, we set the number of un-modeled states $\rvz^m$ according to the number of observed physiological variables to be $M = D/10 = 2, 4\ \mbox{or}\ 8$ (respectively). (We made this choice to reflect the fact that a larger number of physiological variables often necessitates a larger number un-modeled states.) 
The un-modeled states $\rvz^m$ are governed by a nonlinear ODE
\begin{equation*}
\dot{\rvz}_i^m = \text{tanh}(\rmW_1 \rvz_i^m + \rmW_2 \rvz_i^e),
\end{equation*}
with the coefficient matrices $\rmW_1 \in \mathbb{R}^{M \times M}$, $\rmW_2 \in \mathbb{R}^{M \times E}$. For each dataset, we sampled the entries in these matrices independently  from $N(0,1)$.

For each patient $i$, each of the components of  its initial condition $\rvz_i(0)$ were independently drawn from an exponential distribution with rate $\lambda = 100$ (this distribution is also given to the algorithms as the prior distribution). We consider a time horizon of $T = 14$ days; this   is the median length of stay in hospital for Covid-19 patients \cite{rees2020covid}.  

Each patient $i$ will receive a one-time dexamethasone treatment with dosage $d_i$ at some time  $s_i$, where $d_i \sim \text{uniform}[0,10]$ mg and $s_i \sim \text{uniform}[0,T]$. 

The true physiological variables are generated by 
\begin{equation*}
\rvx_i = \rmW_3 \rvz_i + \rmW_4 \rva_i,
\end{equation*}
with the coefficient matrices $\rmW_3 \in \mathbb{R}^{X \times (M+E)}$, $\rmW_4 \in \mathbb{R}^{X \times 1}$.  
For each dataset, each element in these matrices was drawn independently  from $N(0,1)$ and then multiplied by a Bernoulli variable with $p = 0.5$, so that approximately half of the elements in each of these  matrices $\rmW_3$, $\rmW_4$ were 0. 
(We did this in order to reflect the idea that each  physiological variable is only related to some of the latent variables.)  
The measurements are generated by
\begin{equation*}
\rvy_i(t) = \rvx_i(t)  + \epsilon_{it} 
\end{equation*}
with the measurement noise $\epsilon_{it} \sim N(0, \sigma)$ for $\sigma=0.2, 0.4\ \mbox{or}\ 0.8$;  Equation (\ref{eq:error}). 
We first simulate all the daily measurements at $t = 1, 2, \ldots, T$, and then randomly remove measurements with probability $0.5$; this represents the fact that measurements of made irregularly.  

\subsubsection{Hyper-parameter settings}
As a reminder, the number of measured clinical variables is $D$, the number of expert variables is $E$, the number of ML latent variables is $M$. The sample size is $N_0$

The following is the hyperparameter setting used in the simulation study:
\begin{enumerate}
    \item Learning rate: 0.01
    \item Batch size: $\text{min}(50, N_0)$
    \item Early stopping tolerance: 10 epochs
    \item Max iteration: 400
    \item Number of latent variables in NODE: $Z=E+M$, i.e. the true value. (additional settings  $E+M+4$ and $E+M+9$ are reported in the sensitivity analysis)
    \item Number of ML latent variables in LHM: $M$, i.e. the true value.
    \item Latent dimensionality in Encoder: $2D$
    \item Number of layers in NODE: $2$
    \item ODE Solver: adams 
    \item ODE rtol: 1E-7 (library default)
    \item ODE atol: 1E-8 (library default)
\end{enumerate}

\subsubsection{Performance under different lengths of observed history}

\begin{figure}[t!]
  \centering
  \includegraphics[width=1\textwidth]{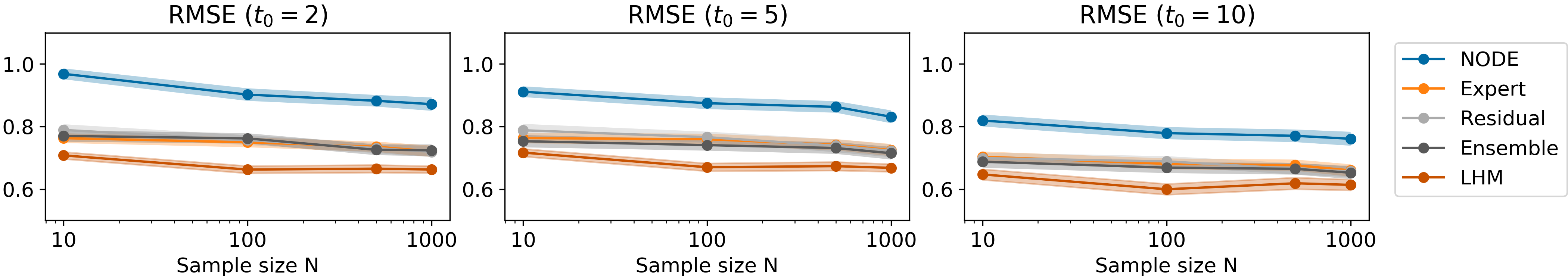}
  \caption{\footnotesize  \textbf{Simulation results under different lengths of observed history $t_0$}. Prediction accuracy on future measurements 
  $\mathcal{Y}[t_0:T]$ given the observed history $\mathcal{Y}[0:t_0]$ as measured by RMSE. The shaded areas represent 95\% confidence intervals.}
  \label{fig:sim_plot_time}
\end{figure}

As a reminder, we use the historical measurements  $\mathcal{Y}[0:t_0]$ up to some time $t_0$ to {\em predict} the future measurements $\mathcal{Y}[t_0:T]$.
To evaluate the performance under different lengths of observed history, we set $t_0 = 2, 5\ \mbox{or} \ 10$ days and use the default setting $\sigma=0.2$ and $M=2$. The results are presented in Figure \ref{fig:sim_plot_time}, where each panel corresponds to a different $t_0$. As expected, the predictive performance improved when longer observed history is available. LHM outperforms the benchmarks for all $t_0$'s we study.

\subsubsection{Performance under different levels of measurement noise}

\begin{figure}[t!]
  \centering
  \includegraphics[width=1\textwidth]{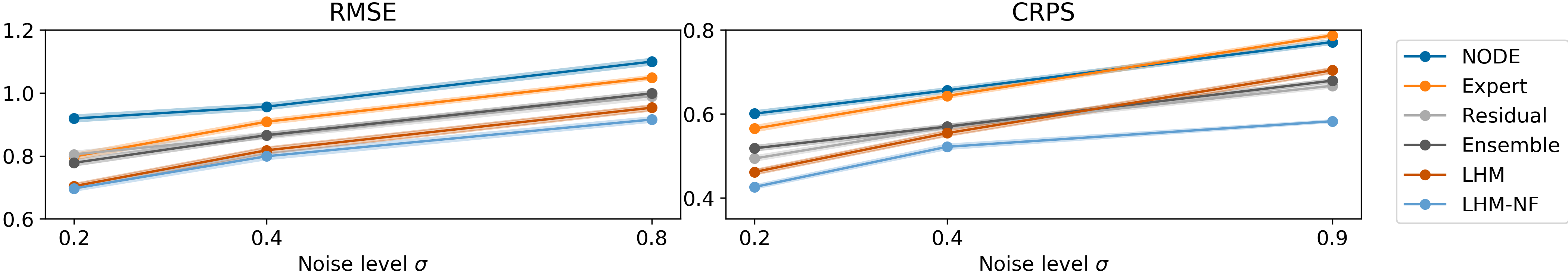}
  \caption{\footnotesize  \textbf{Simulation results under different levels of measurement noise $\sigma$}. Prediction accuracy on future measurements 
  $\mathcal{Y}[5:T]$ given the observed history $\mathcal{Y}[0:5]$ as measured by RMSE and CRPS. The shaded areas represent 95\% confidence intervals.}
  \label{fig:sim_plot_noise}
\end{figure}

In Figure \ref{fig:sim_plot_noise}, we show the model performance under different levels of measurement noise $\sigma = 0.2, 0.4, 0.8$ in a typical simulation with sample size $N_0 = 100$ and $M=2$ un-modeled latent variables $\rvz^m$. In addition to the benchmarks introduced in Section \ref{sec:simulation}, we compared with the LHM using normalizing flow as the variational distribution (LHM-NF) (detailed in the next section). Both LHM and LHM-NF outperform other benchmarks in RMSE. LHM-NF achieves the best CRPS with a fairly big improvement from the second best method (LHM). The result shows LHM's robustness to increased measurement noise.

\subsubsection{Performance gain with Normalizing Flows}

To ensure a fair comparison with existing methods, we use the diagonal Gaussian distribution in LHM as the variational distribution. However, diagonal Gaussian is a restrictive approximation because it does not capture any correlation structure between the latent variables. 

Here we study if using a more flexible distribution will lead to further performance gain. We adopt the planar normalizing flow proposed in \cite{rezende2015variational} with the number of flows set to 4. As is standard in the literature, we amortize the initial conditions $\rvz(0)$ as well as the flow parameters $\rvu$, $\rvw$ and $\rvb$. The following shows a typical simulation with $N_0=100$, $\sigma=0.4$ and $M=2$.

\begin{figure}[t!]
  \centering
  \includegraphics[width=0.7\textwidth]{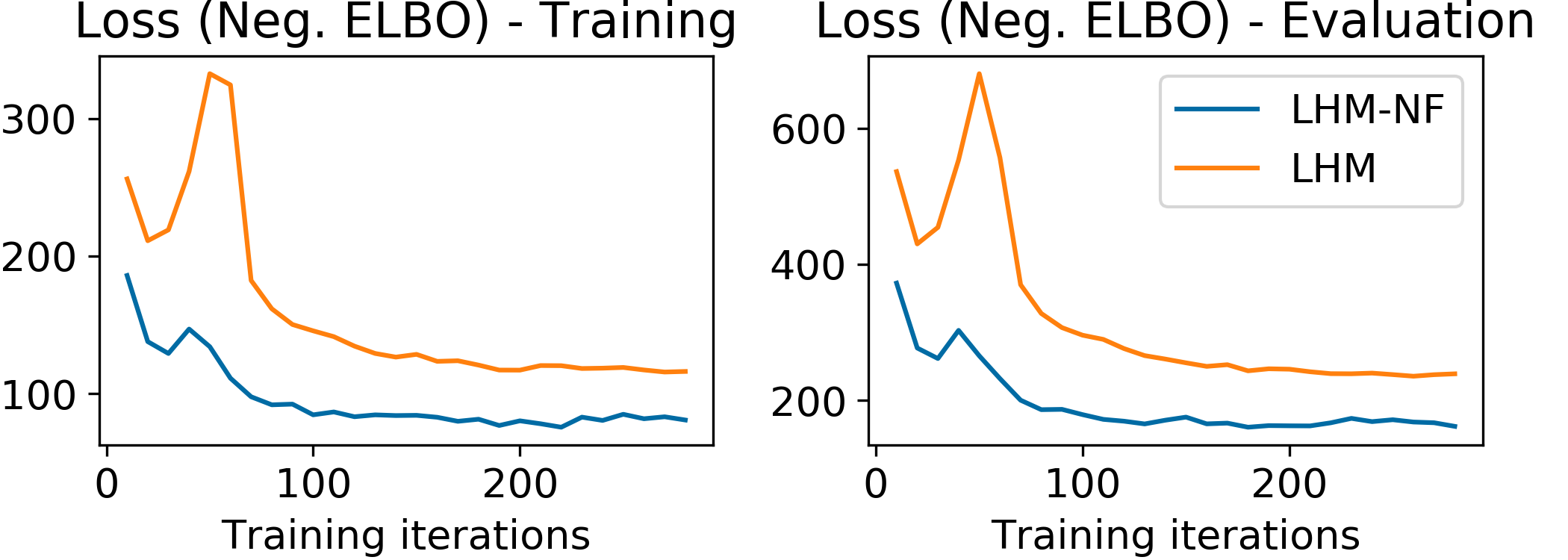}
  \caption{\footnotesize  \textbf{Comparison between the standard LHM and the version with normalizing flow (LHM-NF)}. Loss on training and evaluation datasets are plotted over training iterations. The loss is the negative ELBO. }
  \label{fig:sim_plot_training}
\end{figure}

Figure \ref{fig:sim_plot_training} tracks the loss function (negative ELBO) during training on the training and the evaluation data respectively. As we expected, the version with normalizing flow (LHM-NF) consistently achieves smaller loss on the training data due to the increased flexibility of the variational distribution. The improvement persists when we turn to the evaluation data, and eventually translates into the performance gain illustrated in Figure \ref{fig:sim_plot_noise}. This suggests that using a more flexible variational distribution (e.g. normalizing flow) tends to improve accuracy as well as the uncertainty estimation. 

\subsubsection{Performance of NODE is not sensitive to adding more latent variables}

\begin{figure}[t!]
  \centering
  \includegraphics[width=0.7\textwidth]{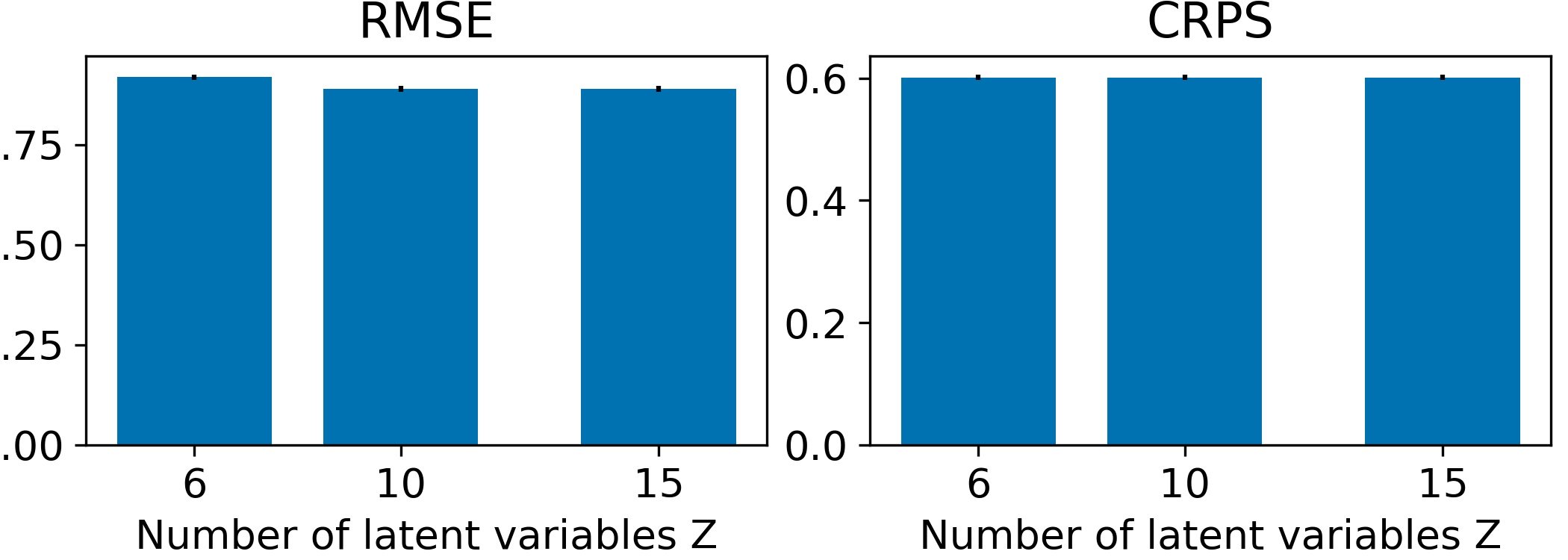}
  \caption{\footnotesize  \textbf{NODE's performance under different numbers of latent variables $Z$}. The data are generated from six \textit{true} latent variables.  Prediction accuracy on future measurements $\mathcal{Y}[5:T]$ given the observed history $\mathcal{Y}[0:5]$ as measured by RMSE and CRPS.}
  \label{fig:sim_plot_latent}
\end{figure}

In the simulations reported above, we set the number of latent variables in NODE to be the true value, i.e. $Z = M + E$. In practice, $Z$ is a hyper-parameter that we do not know a priori. Here we study if the performance is sensitive to the exact choice of $Z$. We consider a setting where the data is generated from $6$ latent variables (including both $\rvz^m$ and $\rvz^e$) and we vary $Z=6, 10, 15$. We present the results in a typical simulation setting with  $N_0=100$, $\sigma=0.2$.
As we show in Figure \ref{fig:sim_plot_latent}, the predictive performance does not significant change even when $Z$ is more than doubled. Note that similar findings have been reported in prior research \cite{dupont2019augmented}. This supports our choice of setting $Z = M + E$ by default.

\subsubsection{Computational resources}

The simulations were performed on a server with a Intel(R) Core(TM) i5-8600K CPU @ 3.60GHz and a Nvidia(R) GeForce(TM) RTX 2080 Ti GPU.
All individual simulations were finished within 3 hours.

\subsection{Extended related works}
\label{app:related_work}

\subsubsection{Hybrid models}

Table \ref{tab:1} categorizes various hybrid modeling frameworks in terms the kind of expert model  and the kind of  machine learning that are used. For illustrative purposes, we  consider a static prediction problem with measurements (covariates) $\rvx\in \mathbb{R}^D$ and target outcome $\rvy\in \mathbb{R}^K$.

\begin{table}[t!]
\footnotesize
\caption{\footnotesize Different methods to create hybrid ML models.  We consider a static prediction problem with covariates $\rvx\in \mathbb{R}^D$ and target outcome $\rvy\in\mathbb{R}^K$ (notations differ from the rest of the paper). The $\rvr:=\rvy-\hat{\rvy}$ denotes the residuals. }
  \label{tab:1}
  \begin{center}
  
\begin{tabular}{@{}lcccc@{}}
\toprule
      \thead{Method}        &  \thead{Example}       & \thead{Expert model} & \thead{ML model} & \thead{Final output}  \\ \midrule
Residual Model       & \cite{liu2019multi,wang2017physics} & $\hat{\rvy}=f^e(\rvx)$   &  $\hat{\rvr}=f^m(\rvx)$              & $\hat{\rvy}+\hat{\rvr}$                \\
Ensemble               & \cite{yao2018tensormol,xu2015data} & $\hat{\rvy}_1=f^e(\rvx)$    & $\hat{\rvy}_2=f^m(\rvx)$              & $w_1\hat{\rvy}_1+w_2\hat{\rvy}_2$                 \\
Feature Extraction & \cite{karpatne2017physics} & $\rvz^e=f^e(\rvx)$ & $\hat{\rvy}=f^m(\rvz^e)$  & $\hat{\rvy}$                \\
LHM        & This work & Eq. \ref{eq:expert}  & Eq. \ref{eq:physiological}, \ref{eq:latent}  & Eq. \ref{eq:physiological} \\ \bottomrule
\end{tabular}
\end{center}
\end{table}

\subsubsection{Other research areas that involve ML and expert ODEs}

There are several research areas that involve ML and expert ODEs but they are \textit{unrelated} to hybrid model or LHM. We briefly describe them for clarification and completeness.

\textbf{Reduced-Order Models.} The expert model may involve a large number of variables, but not all of them are important to the system dynamics (e.g. in large, high-fidelity models of fluid dynamics \cite{lassila2014model}). 
Reduced-Order Models (ROMs) are compact representations of the more complex models \cite{willard2020integrating}. They are often constructed using dimensionality reduction to retain only the most important dynamical characteristics of the original model. ROMs often achieve better estimation efficiency and lower the computational cost. Recently, ML has been applied to ROMs and achieved promising results \cite{chen2012support,xiao2019reduced}. 

In ROMs, we start with an expert model that is \textit{over-complete} and contains redundant variables. In contrast, in LHM, we are given a pharmacological model that is \textit{incomplete}, i.e. it cannot fully explain the high-dimensional clinical measurements or provide the link between expert variables and the measurements.  Hence, LHM is essentially solving the \textit{opposite} problem of ROMs as we are introducing additional machine-learned latent variables into the system. 

\textbf{Using ML to solve expert equations}. Some expert models involves ODEs or PDEs that are computationally challenging to solve (e.g. the  quantum many-body problem \cite{carleo2017solving}). ML has been used to speed up the solution process by making various approximations \cite{han2018solving,sirignano2018dgm}. However, the pharmacological models are generally well-behaved and the standard ODE solvers are able to find the solutions efficiently.

\textbf{Learning unknown ODEs from data} Step-wise regression is a general framework to discover unknown ODEs from data. It applies symbolic or sparse regression to the the observed time derivatives. When these time derivatives are not observed, they are first estimated from the (frequently-sampled) observations (e.g. by finite difference method) \cite{bongard2007automated,brunton2016discovering, rudy2017data}. This approach is not applicable to our setting because the time derivatives of the expert variables $\dot{\rvz}^m$ are not observed or can be easily estimated from the data.

In addition to neural ODEs, Gaussian Processes (GP) have also been used to approximate unknown governing equations \cite{archambeau2007gaussian,ruttor2013approximate}. However, most existing works focus on the discrete-time setting or use fixed step ODE solvers.

\subsubsection{Using Pharmacology/Biology models in ML}

Several other works have proposed to integrate pharmacological models into machine learning. But the problem settings they considered and the approach they took is different from LHM.

\cite{hussain2021neural} introduces a pharmacological model (the log-cell kill model) to modulate the state transition dynamics of a state-space model. Their work considers discrete-time dynamical systems rather than the continuous-time systems we focus on. The authors recognize that the existing log-cell kill models are inadequate to model the disease dynamics (e.g. failure to capture relapses). To address this shortcoming, the authors designed a new set of expert equations to allow for more complex dynamics before integrating them with ML. Hence, this approach requires a deep understanding of the expert model, and a fair amount of mathematical knowledge and manual work to modify the expert model. 
Furthermore, this modification process has to be repeated for a different expert model. 
In contrast, LHM learns the missing dynamics by introducing the ML latent variables $\rvz^m$ and neural ODEs $f^m$.

\cite{yazdani2020systems} considers a problem with more expert variables than observable physiological variables, which is opposite to the setting we consider.\footnote{The Systems Biology models considered in \cite{yazdani2020systems} usually involve a large number of expert variables. This is not the case in the pharmacological models we consider.} 
The problem setting is similar to the reduced-order models discussed above. 
The authors use a neural network with time $t$ as input and outputs the system status at that time. In contrast, LHM uses neural ODEs to model the time derivatives and obtains the system status at time $t$ by solving the ODEs. Finally, the authors evaluate the gradient of the neural network with respect to $t$ by automatic differentiation and introduce an additional loss function to ensure the network gradient matches the expert ODEs. LHM does not involve any heuristic modification on the loss function and follows the standard practice in Bayesian inference. 

\begin{table}[t!]
\footnotesize
\caption{\footnotesize Methods that are related to predicting future outcomes given the intervention. LHM is based on the physical/mechanistic view on causality, which naturally applies to the continuous-time setting. Alternative approaches are developed based on different notions of causality but operate in static or discrete-time settings.}
  \label{tab:causal}
  \begin{center}
\begin{tabular}{llll}
\hline
Framework              & Example & Temporality     & Core component                             \\ \hline
Physical/Mechanistic   & LHM     & Continuous-time & Governing equations (ODEs) of the dynamics \\
Potential outcome (PO) &  \cite{rubin2005causal}  & Static / discrete & Statistical properties of the PO           \\
Causal graphical       & \cite{pearl1995causal} & Static / discrete & Causal graphs (DAG) of the variables       \\
Causal structural      & \cite{pearl2009causality} & Static / discrete & Structural equations between the variables  \\ \hline
\end{tabular}
\end{center}
\end{table}

\subsubsection{Causal treatment effect estimation}

LHM predicts the disease progression under treatments based on governing ODEs. 
The causal inference literature studies related problems that involve predicting/estimating treatment effects.
Although causal inference is a diverse field, most existing methods operate in the static or the discrete-time setting. In contrast, LHM operates in the continuous-time setting due to the irregularity of clinical measurements.
Here we compare the approach taken by LHM with other approaches in the literature for completeness. These methods are summarized in Table \ref{tab:causal}.

As discussed in Section \ref{sec:related_work}, LHM predicts the future health status given treatments using the governing equations (ODEs). This corresponds to the mechanistic (or physical) notion of the causality, which is recognized as the ``gold standard'' for modeling natural phenomena by \cite{scholkopf2021toward}. The governing ODEs describes the system dynamics in \textit{continuous time}, which is essential because the clinical measurements are made at irregular time points. 

The potential outcome framework widely used in Statistics is based on a different notion of causality \cite{rubin2005causal}. It makes assumptions about the statistical properties of the unobservable potential outcomes (e.g. independence) to make inference about the (conditional) average treatment effect. Here, the focus is not on using or discovering the underlying governing equations, but on leveraging the statistical associations between the observed and the potential outcomes. Unlike the mechanistic framework, the potential outcome framework does not require the system to be observed over time, making it suitable for problems involving only static variables.

The causal graphical model \cite{pearl1995causal} uses another notion of causality. It describes the causal structure between variables as a graph (typically a directed acyclic graph, DAG). Various identification strategies have been developed to infer the causal effect given the graph (e.g. the backdoor criterion \cite{pearl2009causality}). 
A closely related framework is the structural causal model \cite{pearl2009causality}, where a set of structural equations are given in addition to the causal graph. Typically, the structural equations are standard equations that link the (static or discretely sampled) variables, but they are not ODEs that describe the continuous-time dynamics.

\subsection{Real data experiment}
\label{app:real_data}

\subsubsection{List of clinical variables}

We use the measurements of the following temporal physiological variables. These variables are chosen by our clinical collaborators and reflect the information accessible and important to a clinician when deciding the treatment plan. They  include vital signals, lung mechanics, and the biomarkers measured in blood tests. 
\begin{enumerate}
    \item P/F ratio
    \item PEEP
    \item SOFA
    \item Temperature
    \item Arterial blood pressure
    \item Heart Rate
    \item Bilirubin
    \item Thrombocytes
    \item Leukocytes
    \item Creatinine
    \item C Reactive Protein
    \item Arterial lactate
    \item Creatine kinase
    \item Glucose
    \item Alanine transaminase
    \item Aspartate transaminase
    \item Prone positioning
    \item Tidal volume
    \item Driving pressure
    \item FiO2
    \item Lung compliance (static)
    \item Respiratory rate
    \item Pressure above PEEP
    \item Arterial PaCO2
    \item Arterial PH
    \item PaCO2 (unspecified)
    \item PH (unspecified)
\end{enumerate}

We used the following static covariates:
\begin{enumerate}
    \item Age
    \item Sex
    \item Body Mass Index
    \item Comorbidity: cirrhosis
    \item Comorbidity: chronic dialysis
    \item Comorbidity: chronic renal insufficiency
    \item Comorbidity: diabetes
    \item Comorbidity: cardiovascular insufficiency
    \item Comorbidity: copd
    \item Comorbidity: respiratory insufficiency
    \item Comorbidity: immunodeficiency
\end{enumerate}

\subsubsection{Eligibility criterion}

We selected all patients in DDW who stayed in the ICU for more than 2 days and less than 31 days (2097 out of 3464). Patients with a very short length of stay will not give us enough data points for training or evaluation. 

\subsubsection{Hyper-parameter settings}

The following is the hyperparameter setting used in the real-data study. They are decided based on a pilot study. 
\begin{enumerate}
    \item Learning rate: 0.01
    \item Batch size: $100$
    \item Early stopping tolerance: 10 epochs
    \item Max iteration: 1500
    \item Number of latent variables in NODE: $20$
    \item Number of ML latent variables in LHM: $15$ (this is to ensure the total number of latent variables is the same as NODE).
    \item Latent dimensionality in Encoder: $1.2D$
    \item Number of layers in NODE: $2$
    \item ODE Solver: adams
    \item ODE rtol: 1E-7 (library default)
    \item ODE atol: 1E-8 (library default)
\end{enumerate}

\subsubsection{Accuracy over different time horizons}

Table \ref{real-table-horizon} shows the performance over different prediction horizons $H$ given $N_0=1000$ training samples. LHM achieves the best or equally the best performance in all cases.

\subsubsection{License and anonymity }
Access to the DDW is regulated. We have signed an end user license before access to the data was granted. All data were pseudonymized in DDW.

\begin{table}[t]
  \caption{Prediction accuracy (RMSE) on $\mathcal{Y}[t_0:t_0+H]$ over different time horizons $H$ (hours). The standard deviations are shown in the brackets.}
  \label{real-table-horizon}
  \footnotesize
  \centering
\begin{tabular}{@{}lcccc@{}}
\toprule
Method \textbackslash H= & 6                     & 12                    & 24                    & 72                    \\ \midrule
Expert                   & 0.734 (0.99)          & 0.724 (1.00)          & 0.713 (0.03)          & 0.993 (0.03)          \\
Residual                 & 0.555 (0.98)          & 0.575 (1.08)          & 0.607 (0.04)          & 0.983 (0.05)          \\
Ensemble                 & 0.556 (0.71)          & 0.573 (0.73)          & 0.599 (0.04)          & \textbf{0.713 (0.05)} \\
NODE                     & 0.661 (1.00)          & 0.654 (1.00)          & 0.650 (0.02)          & 0.996 (0.02)          \\
ODE2VAE                  & 0.627 (1.11)          & 0.616 (1.09)          & 0.619 (0.02)          & 1.113 (0.01)         \\
GRU-ODE                  & 0.549 (0.71)          & 0.571 (0.72)          & 0.601 (0.04)          & \textbf{0.711 (0.05)} \\
Time LSTM                & 0.610 (0.81)          & 0.620 (0.82)          & 0.631 (0.04)          & 0.807 (0.05)          \\
LHM                      & \textbf{0.517 (0.72)} & \textbf{0.511 (0.73)} & \textbf{0.511 (0.03)} & \textbf{0.691 (0.03)} \\ \bottomrule
\end{tabular}
\end{table}

\subsection{Practical extensions}
\label{app:extensions}

\textbf{Incorporating static covariates.} Static covariates such as the demographics often impact disease progression. We can easily incorporate these variables in LHM by treating them as time-constant ``treatments''. This will allow the static covariates to impact the latent dynamics as well as the mapping between the latent and physiological variables (Equation \ref{eq:expert} to \ref{eq:latent}).

\textbf{Informative sampling.} It is well known that the sampling frequency may carry information about the variables being measured (e.g. clinicians tend to take measurements more often if a patient is critically ill) \cite{alaa2017learning}. One approach to incorporate informative sampling is to explicitly model it as a marked point process \cite{rubanova2019latent}. Another popular approach is to concatenate the measurements $\rvx$ with the masking vector that indicates which variable is measured, and train the model on the extended measurement vector \cite{kidger2020neural}. Both approaches are compatible with LHM.

\textbf{Correcting model mis-specification.} \textit{Equation Replacement} is a general approach that applies to any \textit{misspecified} expert model and it can be combined with all the methods discussed above, and to   LHM \cite{hamilton2017hybrid,parish2016paradigm,zhang2018real}.
In this approach, one first identifies which equations in the expert model are misspecified, and then replaces these by flexible function approximators (such as neural networks), that will approximate the true equation after training. Equation replacement only attempts to correct the misspecifications in the original model, but does not introduce any new variables. 

\textbf{Efficient online inference.} The inference method presented in Section \ref{sec:inference} requires to re-process the entire history each time a new measurement is made. Instead, it may be desirable to incrementally update the posterior of $\rvz_i(0)$ based on the most recent measurement only. Fortunately, online Bayesian update (also known as Bayesian Filtering) is a well studied problem with many proven solutions (e.g. Kalman filter and extensions \cite{ribeiro2004kalman}). These inference methods can be applied when the efficiency of online inference is of concern.

\textbf{Improving encoder architecture.} For a fair comparison with related works, we used the reversed time-aware LSTM encoder proposed in \cite{chen2018neural}. Essentially, it is a LSTM with the observation time as an additional input channel and running backward through time. To further improve performance, one may explore other architectures. Essentially, any architecture that takes irregularly sampled data as input is applicable. Examples include the Neural Controlled Ordinary Differential Equation \cite{kidger2020neural} and the Neural ODE Processes \cite{norcliffe2021neural}. 

\bibliographystyle{plain}
\bibliography{preprint_version}

\begin{thebibliography}{10}

\bibitem{abbas2014cellular}
Abul~K Abbas, Andrew~H Lichtman, and Shiv Pillai.
\newblock {\em Cellular and molecular immunology E-book}.
\newblock Elsevier Health Sciences, 2014.

\bibitem{agoram2007role}
Balaji~M Agoram, Steven~W Martin, and Piet~H van~der Graaf.
\newblock The role of mechanism-based pharmacokinetic--pharmacodynamic (pk--pd)
  modelling in translational research of biologics.
\newblock {\em Drug discovery today}, 12(23-24):1018--1024, 2007.

\bibitem{alaa2017learning}
Ahmed~M Alaa, Scott Hu, and Mihaela Schaar.
\newblock Learning from clinical judgments: Semi-markov-modulated marked hawkes
  processes for risk prognosis.
\newblock In {\em International Conference on Machine Learning}, pages 60--69.
  PMLR, 2017.

\bibitem{alaaattentive}
Ahmed~M Alaa and Mihaela van~der Schaar.
\newblock Attentive state-space modeling of disease progression.
\newblock In {\em 33rd Conference on Neural Information Processing Systems
  (NeurIPS 2019)}, 2019.

\bibitem{archambeau2007gaussian}
Cedric Archambeau, Dan Cornford, Manfred Opper, and John Shawe-Taylor.
\newblock Gaussian process approximations of stochastic differential equations.
\newblock In {\em Gaussian Processes in Practice}, pages 1--16. PMLR, 2007.

\bibitem{aronson2017biomarkers}
Jeffrey~K Aronson and Robin~E Ferner.
\newblock Biomarkers—a general review.
\newblock {\em Current protocols in pharmacology}, 76(1):9--23, 2017.

\bibitem{baytas2017patient}
Inci~M Baytas, Cao Xiao, Xi~Zhang, Fei Wang, Anil~K Jain, and Jiayu Zhou.
\newblock Patient subtyping via time-aware lstm networks.
\newblock In {\em Proceedings of the 23rd ACM SIGKDD international conference
  on knowledge discovery and data mining}, pages 65--74, 2017.

\bibitem{bertalan2019learning}
Tom Bertalan, Felix Dietrich, Igor Mezi{\'c}, and Ioannis~G Kevrekidis.
\newblock On learning hamiltonian systems from data.
\newblock {\em Chaos: An Interdisciplinary Journal of Nonlinear Science},
  29(12):121107, 2019.

\bibitem{bongard2007automated}
Josh Bongard and Hod Lipson.
\newblock Automated reverse engineering of nonlinear dynamical systems.
\newblock {\em Proceedings of the National Academy of Sciences},
  104(24):9943--9948, 2007.

\bibitem{brunton2016discovering}
Steven~L Brunton, Joshua~L Proctor, and J~Nathan Kutz.
\newblock Discovering governing equations from data by sparse identification of
  nonlinear dynamical systems.
\newblock {\em Proceedings of the national academy of sciences},
  113(15):3932--3937, 2016.

\bibitem{carleo2017solving}
Giuseppe Carleo and Matthias Troyer.
\newblock Solving the quantum many-body problem with artificial neural
  networks.
\newblock {\em Science}, 355(6325):602--606, 2017.

\bibitem{chen2012support}
Gang Chen, Yingtao Zuo, Jian Sun, and Yueming Li.
\newblock Support-vector-machine-based reduced-order model for limit cycle
  oscillation prediction of nonlinear aeroelastic system.
\newblock {\em Mathematical problems in engineering}, 2012, 2012.

\bibitem{chen2018neural}
Ricky~TQ Chen, Yulia Rubanova, Jesse Bettencourt, and David Duvenaud.
\newblock Neural ordinary differential equations.
\newblock {\em arXiv preprint arXiv:1806.07366}, 2018.

\bibitem{cho2014learning}
Kyunghyun Cho, Bart Van~Merri{\"e}nboer, Caglar Gulcehre, Dzmitry Bahdanau,
  Fethi Bougares, Holger Schwenk, and Yoshua Bengio.
\newblock Learning phrase representations using rnn encoder-decoder for
  statistical machine translation.
\newblock {\em CoRR, abs/1406.1078}, 2014.

\bibitem{choi2016retain}
Edward Choi, Mohammad~Taha Bahadori, Joshua~A Kulas, Andy Schuetz, Walter~F
  Stewart, and Jimeng Sun.
\newblock Retain: An interpretable predictive model for healthcare using
  reverse time attention mechanism.
\newblock {\em Advances in Neural Information Processing Systems}, pages
  3512--3520, 2016.

\bibitem{cox2020co}
Michael~J Cox, Nicholas Loman, Debby Bogaert, and Justin O'Grady.
\newblock Co-infections: potentially lethal and unexplored in covid-19.
\newblock {\em The Lancet Microbe}, 1(1):e11, 2020.

\bibitem{dai2021prototype}
Wei Dai, Rohit Rao, Anna Sher, Nessy Tania, Cynthia~J Musante, and Richard
  Allen.
\newblock A prototype qsp model of the immune response to sars-cov-2 for
  community development.
\newblock {\em CPT: pharmacometrics \& systems pharmacology}, 10(1):18--29,
  2021.

\bibitem{danhof2016systems}
Meindert Danhof.
\newblock Systems pharmacology--towards the modeling of network interactions.
\newblock {\em European Journal of Pharmaceutical Sciences}, 94:4--14, 2016.

\bibitem{danhof2007mechanism}
Meindert Danhof, Joost de~Jongh, Elizabeth~CM De~Lange, Oscar Della~Pasqua,
  Bart~A Ploeger, and Rob~A Voskuyl.
\newblock Mechanism-based pharmacokinetic-pharmacodynamic modeling: biophase
  distribution, receptor theory, and dynamical systems analysis.
\newblock {\em Annu. Rev. Pharmacol. Toxicol.}, 47:357--400, 2007.

\bibitem{de2019gru}
Edward De~Brouwer, Jaak Simm, Adam Arany, and Yves Moreau.
\newblock Gru-ode-bayes: Continuous modeling of sporadically-observed time
  series.
\newblock In {\em 33rd Conference on Neural Information Processing Systems
  (NeurIPS 2019)}, 2019.

\bibitem{dupont2019augmented}
Emilien Dupont, Arnaud Doucet, and Yee~Whye Teh.
\newblock Augmented neural odes.
\newblock In {\em 33rd Conference on Neural Information Processing Systems
  (NeurIPS 2019)}, 2019.

\bibitem{earp2008pharmacokinetics}
Justin~C Earp, Nancy~A Pyszczynski, Diana~S Molano, and William~J Jusko.
\newblock Pharmacokinetics of dexamethasone in a rat model of rheumatoid
  arthritis.
\newblock {\em Biopharmaceutics \& drug disposition}, 29(6):366--372, 2008.

\bibitem{fajgenbaum2020cytokine}
David~C Fajgenbaum and Carl~H June.
\newblock Cytokine storm.
\newblock {\em New England Journal of Medicine}, 383(23):2255--2273, 2020.

\bibitem{falvey2015disease}
James~D Falvey, Teagan Hoskin, Berrie Meijer, Anna Ashcroft, Russell Walmsley,
  Andrew~S Day, and Richard~B Gearry.
\newblock Disease activity assessment in ibd: clinical indices and biomarkers
  fail to predict endoscopic remission.
\newblock {\em Inflammatory bowel diseases}, 21(4):824--831, 2015.

\bibitem{fioretto2020predicting}
Ferdinando Fioretto, Terrence~WK Mak, and Pascal Van~Hentenryck.
\newblock Predicting ac optimal power flows: Combining deep learning and
  lagrangian dual methods.
\newblock In {\em Proceedings of the AAAI Conference on Artificial
  Intelligence}, volume 34 01, pages 630--637, 2020.

\bibitem{fleuren2021large}
Lucas~M Fleuren, Daan~P de~Bruin, Michele Tonutti, Robbert~CA Lalisang, and
  Paul~WG Elbers.
\newblock Large-scale icu data sharing for global collaboration: the first 1633
  critically ill covid-19 patients in the dutch data warehouse.
\newblock {\em Intensive care medicine}, 47(4):478--481, 2021.

\bibitem{fleuren2020machine}
Lucas~M Fleuren, Patrick Thoral, Duncan Shillan, Ari Ercole, Paul~WG Elbers,
  and Right Data Right Now Collaborators Mark Hoogendoorn Ben Gibbison Thomas
  LT Klausch Tingjie Guo Luca F. Roggeveen Eleonora L. Swart Armand~RJ Girbes.
\newblock Machine learning in intensive care medicine: ready for take-off?
\newblock {\em Intensive care medicine}, 46:1486--1488, 2020.

\bibitem{frank2003clinical}
Richard Frank and Richard Hargreaves.
\newblock Clinical biomarkers in drug discovery and development.
\newblock {\em Nature reviews Drug discovery}, 2(7):566--580, 2003.

\bibitem{gesztelyi2012hill}
Rudolf Gesztelyi, Judit Zsuga, Adam Kemeny-Beke, Balazs Varga, Bela Juhasz, and
  Arpad Tosaki.
\newblock The hill equation and the origin of quantitative pharmacology.
\newblock {\em Archive for history of exact sciences}, 66(4):427--438, 2012.

\bibitem{gilchrist2002modeling}
Michael~A Gilchrist and Akira Sasaki.
\newblock Modeling host--parasite coevolution: a nested approach based on
  mechanistic models.
\newblock {\em Journal of Theoretical Biology}, 218(3):289--308, 2002.

\bibitem{greydanus2019hamiltonian}
Sam Greydanus, Misko Dzamba, and Jason Yosinski.
\newblock Hamiltonian neural networks.
\newblock In {\em 33rd Conference on Neural Information Processing Systems
  (NeurIPS 2019)}, 2019.

\bibitem{recovery2021dexamethasone}
RECOVERY~Collaborative Group.
\newblock Dexamethasone in hospitalized patients with covid-19.
\newblock {\em New England Journal of Medicine}, 384(8):693--704, 2021.

\bibitem{guen2020disentangling}
Vincent~Le Guen and Nicolas Thome.
\newblock Disentangling physical dynamics from unknown factors for unsupervised
  video prediction.
\newblock In {\em Proceedings of the IEEE/CVF Conference on Computer Vision and
  Pattern Recognition}, pages 11474--11484, 2020.

\bibitem{hamilton2017hybrid}
Franz Hamilton, Alun~L Lloyd, and Kevin~B Flores.
\newblock Hybrid modeling and prediction of dynamical systems.
\newblock {\em PLoS computational biology}, 13(7):e1005655, 2017.

\bibitem{han2018solving}
Jiequn Han, Arnulf Jentzen, and E~Weinan.
\newblock Solving high-dimensional partial differential equations using deep
  learning.
\newblock {\em Proceedings of the National Academy of Sciences},
  115(34):8505--8510, 2018.

\bibitem{higgins2016beta}
Irina Higgins, Loic Matthey, Arka Pal, Christopher Burgess, Xavier Glorot,
  Matthew Botvinick, Shakir Mohamed, and Alexander Lerchner.
\newblock beta-vae: Learning basic visual concepts with a constrained
  variational framework.
\newblock {\em ICLR}, 2017.

\bibitem{holford2013pharmacokinetic}
Nick Holford, Young-A Heo, and Brian Anderson.
\newblock A pharmacokinetic standard for babies and adults.
\newblock {\em Journal of pharmaceutical sciences}, 102(9):2941--2952, 2013.

\bibitem{hussain2021neural}
Zeshan Hussain, Rahul~G Krishnan, and David Sontag.
\newblock Neural pharmacodynamic state space modeling.
\newblock {\em arXiv preprint arXiv:2102.11218}, 2021.

\bibitem{johnson2016mimic}
Alistair~EW Johnson, Tom~J Pollard, Lu~Shen, H~Lehman Li-Wei, Mengling Feng,
  Mohammad Ghassemi, Benjamin Moody, Peter Szolovits, Leo~Anthony Celi, and
  Roger~G Mark.
\newblock Mimic-iii, a freely accessible critical care database.
\newblock {\em Scientific data}, 3(1):1--9, 2016.

\bibitem{karpatne2017physics}
Anuj Karpatne, William Watkins, Jordan Read, and Vipin Kumar.
\newblock Physics-guided neural networks (pgnn): An application in lake
  temperature modeling.
\newblock {\em arXiv preprint arXiv:1710.11431}, 2017.

\bibitem{katzung2012basic}
Bertram~G Katzung.
\newblock {\em Basic and clinical pharmacology}.
\newblock Mc Graw Hill, 2012.

\bibitem{keller2020effect}
Marla~J Keller, Elizabeth~A Kitsis, Shitij Arora, Jen-Ting Chen, Shivani
  Agarwal, Michael~J Ross, Yaron Tomer, and William Southern.
\newblock Effect of systemic glucocorticoids on mortality or mechanical
  ventilation in patients with covid-19.
\newblock {\em Journal of hospital medicine}, 15(8):489--493, 2020.

\bibitem{khader2020targeting}
Shabaana~A Khader, Maziar Divangahi, Willem Hanekom, Philip~C Hill, Markus
  Maeurer, Karen~W Makar, Katrin~D Mayer-Barber, Musa~M Mhlanga, Elisa Nemes,
  Larry~S Schlesinger, et~al.
\newblock Targeting innate immunity for tuberculosis vaccination.
\newblock {\em The Journal of clinical investigation}, 129(9):3482--3491, 2020.

\bibitem{kidger2020neural}
Patrick Kidger, James Morrill, James Foster, and Terry Lyons.
\newblock Neural controlled differential equations for irregular time series.
\newblock In {\em 34th Conference on Neural Information Processing Systems
  (NeurIPS 2020)}, 2020.

\bibitem{kingma2014adam}
Diederik~P Kingma and Jimmy Ba.
\newblock Adam: A method for stochastic optimization.
\newblock {\em arXiv preprint arXiv:1412.6980}, 2014.

\bibitem{kingma2013auto}
Diederik~P Kingma and Max Welling.
\newblock Auto-encoding variational bayes.
\newblock {\em arXiv preprint arXiv:1312.6114}, 2013.

\bibitem{koehler2020defining}
Philipp Koehler, Matteo Bassetti, Arunaloke Chakrabarti, Sharon~CA Chen,
  Arnaldo~Lopes Colombo, Martin Hoenigl, Nikolay Klimko, Cornelia
  Lass-Fl{\"o}rl, Rita~O Oladele, Donald~C Vinh, et~al.
\newblock Defining and managing covid-19-associated pulmonary aspergillosis:
  the 2020 ecmm/isham consensus criteria for research and clinical guidance.
\newblock {\em The Lancet Infectious Diseases}, 2020.

\bibitem{kramer1974pharmacokinetics}
William~G Kramer, Richard~P Lewis, Tyson~C Cobb, Wilbur~F Forester, James~A
  Visconti, Lee~A Wanke, Harold~G Boxenbaum, and Richard~H Reuning.
\newblock Pharmacokinetics of digoxin: comparison of a two-and a
  three-compartment model in man.
\newblock {\em Journal of pharmacokinetics and biopharmaceutics},
  2(4):299--312, 1974.

\bibitem{lassila2014model}
Toni Lassila, Andrea Manzoni, Alfio Quarteroni, and Gianluigi Rozza.
\newblock Model order reduction in fluid dynamics: challenges and perspectives.
\newblock {\em Reduced Order Methods for modeling and computational reduction},
  pages 235--273, 2014.

\bibitem{lee2018determining}
Michael~D Lee and Wolf Vanpaemel.
\newblock Determining informative priors for cognitive models.
\newblock {\em Psychonomic Bulletin \& Review}, 25(1):114--127, 2018.

\bibitem{liu2019multi}
Dehao Liu and Yan Wang.
\newblock Multi-fidelity physics-constrained neural network and its application
  in materials modeling.
\newblock {\em Journal of Mechanical Design}, 141(12), 2019.

\bibitem{metzler1971usefulness}
Carl~M Metzler.
\newblock Usefulness of the two-compartment open model in pharmacokinetics.
\newblock {\em Journal of the American Statistical Association},
  66(333):49--53, 1971.

\bibitem{moore2020cytokine}
John~B Moore and Carl~H June.
\newblock Cytokine release syndrome in severe covid-19.
\newblock {\em Science}, 368(6490):473--474, 2020.

\bibitem{mortensen2001c}
Richard~F Mortensen.
\newblock C-reactive protein, inflammation, and innate immunity.
\newblock {\em Immunologic research}, 24(2):163--176, 2001.

\bibitem{muralidhar2018incorporating}
Nikhil Muralidhar, Mohammad~Raihanul Islam, Manish Marwah, Anuj Karpatne, and
  Naren Ramakrishnan.
\newblock Incorporating prior domain knowledge into deep neural networks.
\newblock In {\em 2018 IEEE International Conference on Big Data (Big Data)},
  pages 36--45. IEEE, 2018.

\bibitem{guideline}
NIH.
\newblock Therapeutic management of adults with covid-19.
\newblock
  \url{https://www.covid19treatmentguidelines.nih.gov/therapeutic-management/}.
\newblock Accessed: 2021-05-26.

\bibitem{norcliffe2021neural}
Alexander Norcliffe, Cristian Bodnar, Ben Day, Jacob Moss, and Pietro Lio.
\newblock Neural ode processes.
\newblock {\em ICLR}, 2021.

\bibitem{pakman2020neural}
Ari Pakman, Yueqi Wang, Catalin Mitelut, JinHyung Lee, and Liam Paninski.
\newblock Neural clustering processes.
\newblock In {\em International Conference on Machine Learning}, pages
  7455--7465. PMLR, 2020.

\bibitem{parascandolo2018learning}
Giambattista Parascandolo, Niki Kilbertus, Mateo Rojas-Carulla, and Bernhard
  Sch{\"o}lkopf.
\newblock Learning independent causal mechanisms.
\newblock In {\em International Conference on Machine Learning}, pages
  4036--4044. PMLR, 2018.

\bibitem{parish2016paradigm}
Eric~J Parish and Karthik Duraisamy.
\newblock A paradigm for data-driven predictive modeling using field inversion
  and machine learning.
\newblock {\em Journal of Computational Physics}, 305:758--774, 2016.

\bibitem{pearl1995causal}
Judea Pearl.
\newblock Causal diagrams for empirical research.
\newblock {\em Biometrika}, 82(4):669--688, 1995.

\bibitem{pearl2009causality}
Judea Pearl.
\newblock {\em Causality}.
\newblock Cambridge university press, 2009.

\bibitem{perko2013differential}
Lawrence Perko.
\newblock {\em Differential equations and dynamical systems}, volume~7.
\newblock Springer Science \& Business Media, 2013.

\bibitem{read2019process}
Jordan~S Read, Xiaowei Jia, Jared Willard, Alison~P Appling, Jacob~A Zwart,
  Samantha~K Oliver, Anuj Karpatne, Gretchen~JA Hansen, Paul~C Hanson, William
  Watkins, et~al.
\newblock Process-guided deep learning predictions of lake water temperature.
\newblock {\em Water Resources Research}, 55(11):9173--9190, 2019.

\bibitem{rees2020covid}
Eleanor~M Rees, Emily~S Nightingale, Yalda Jafari, Naomi~R Waterlow, Samuel
  Clifford, Carl~AB Pearson, Thibaut Jombart, Simon~R Procter, Gwenan~M Knight,
  CMMID~Working Group, et~al.
\newblock Covid-19 length of hospital stay: a systematic review and data
  synthesis.
\newblock {\em BMC medicine}, 18(1):1--22, 2020.

\bibitem{rezende2015variational}
Danilo Rezende and Shakir Mohamed.
\newblock Variational inference with normalizing flows.
\newblock In {\em International Conference on Machine Learning}, pages
  1530--1538. PMLR, 2015.

\bibitem{ribeiro2004kalman}
Maria~Isabel Ribeiro.
\newblock Kalman and extended kalman filters: Concept, derivation and
  properties.
\newblock {\em Institute for Systems and Robotics}, 43:46, 2004.

\bibitem{rubanova2019latent}
Yulia Rubanova, Ricky~TQ Chen, and David Duvenaud.
\newblock Latent ordinary differential equations for irregularly-sampled time
  series.
\newblock In {\em 33th Conference on Neural Information Processing Systems
  (NeurIPS 2019)}, 2019.

\bibitem{rubin2005causal}
Donald~B Rubin.
\newblock Causal inference using potential outcomes: Design, modeling,
  decisions.
\newblock {\em Journal of the American Statistical Association},
  100(469):322--331, 2005.

\bibitem{rudy2017data}
Samuel~H Rudy, Steven~L Brunton, Joshua~L Proctor, and J~Nathan Kutz.
\newblock Data-driven discovery of partial differential equations.
\newblock {\em Science Advances}, 3(4):e1602614, 2017.

\bibitem{ruttor2013approximate}
Andreas Ruttor, Philipp Batz, and Manfred Opper.
\newblock Approximate gaussian process inference for the drift function in
  stochastic differential equations.
\newblock In {\em Advances in Neural Information Processing Systems}, pages
  2040--2048. Citeseer, 2013.

\bibitem{scholkopf2021toward}
Bernhard Sch{\"o}lkopf, Francesco Locatello, Stefan Bauer, Nan~Rosemary Ke, Nal
  Kalchbrenner, Anirudh Goyal, and Yoshua Bengio.
\newblock Toward causal representation learning.
\newblock {\em Proceedings of the IEEE}, 109(5):612--634, 2021.

\bibitem{schutt2017schnet}
Kristof~T Sch{\"u}tt, Pieter-Jan Kindermans, Huziel~E Sauceda, Stefan Chmiela,
  Alexandre Tkatchenko, and Klaus-Robert M{\"u}ller.
\newblock Schnet: A continuous-filter convolutional neural network for modeling
  quantum interactions.
\newblock {\em arXiv preprint arXiv:1706.08566}, 2017.

\bibitem{shillan2019use}
Duncan Shillan, Jonathan~AC Sterne, Alan Champneys, and Ben Gibbison.
\newblock Use of machine learning to analyse routinely collected intensive care
  unit data: a systematic review.
\newblock {\em Critical Care}, 23(1):1--11, 2019.

\bibitem{sirignano2018dgm}
Justin Sirignano and Konstantinos Spiliopoulos.
\newblock Dgm: A deep learning algorithm for solving partial differential
  equations.
\newblock {\em Journal of computational physics}, 375:1339--1364, 2018.

\bibitem{spoorenberg2014pharmacokinetics}
Simone~MC Spoorenberg, Vera~HM Deneer, Jan~C Grutters, Astrid~E Pulles,
  GP~Voorn, Ger~T Rijkers, Willem Jan~W Bos, and Ewoudt~MW van~de Garde.
\newblock Pharmacokinetics of oral vs. intravenous dexamethasone in patients
  hospitalized with community-acquired pneumonia.
\newblock {\em British journal of clinical pharmacology}, 78(1):78--83, 2014.

\bibitem{stronge2018impact}
William~James Stronge.
\newblock {\em Impact mechanics}.
\newblock Cambridge university press, 2018.

\bibitem{thomas2018tensor}
Nathaniel Thomas, Tess Smidt, Steven Kearnes, Lusann Yang, Li~Li, Kai Kohlhoff,
  and Patrick Riley.
\newblock Tensor field networks: Rotation-and translation-equivariant neural
  networks for 3d point clouds.
\newblock {\em arXiv preprint arXiv:1802.08219}, 2018.

\bibitem{tomazini2020effect}
Bruno~M Tomazini, Israel~S Maia, Alexandre~B Cavalcanti, Otavio Berwanger,
  Regis~G Rosa, Viviane~C Veiga, Alvaro Avezum, Renato~D Lopes, Flavia~R Bueno,
  Maria Vitoria~AO Silva, et~al.
\newblock Effect of dexamethasone on days alive and ventilator-free in patients
  with moderate or severe acute respiratory distress syndrome and covid-19: the
  codex randomized clinical trial.
\newblock {\em Jama}, 324(13):1307--1316, 2020.

\bibitem{van2020corticosteroid}
Judith van Paassen, Jeroen~S Vos, Eva~M Hoekstra, Katinka~MI Neumann, Pauline~C
  Boot, and Sesmu~M Arbous.
\newblock Corticosteroid use in covid-19 patients: a systematic review and
  meta-analysis on clinical outcomes.
\newblock {\em Critical Care}, 24(1):1--22, 2020.

\bibitem{wang2017physics}
Jian-Xun Wang, Jin-Long Wu, and Heng Xiao.
\newblock Physics-informed machine learning approach for reconstructing
  reynolds stress modeling discrepancies based on dns data.
\newblock {\em Physical Review Fluids}, 2(3):034603, 2017.

\bibitem{wang2014unsupervised}
Xiang Wang, David Sontag, and Fei Wang.
\newblock Unsupervised learning of disease progression models.
\newblock In {\em Proceedings of the 20th ACM SIGKDD international conference
  on Knowledge discovery and data mining}, pages 85--94, 2014.

\bibitem{willard2020integrating}
Jared Willard, Xiaowei Jia, Shaoming Xu, Michael Steinbach, and Vipin Kumar.
\newblock Integrating physics-based modeling with machine learning: A survey.
\newblock {\em arXiv preprint arXiv:2003.04919}, 2020.

\bibitem{wu2018physics}
Jin-Long Wu, Heng Xiao, and Eric Paterson.
\newblock Physics-informed machine learning approach for augmenting turbulence
  models: A comprehensive framework.
\newblock {\em Physical Review Fluids}, 3(7):074602, 2018.

\bibitem{xiao2019reduced}
D~Xiao, CE~Heaney, L~Mottet, F~Fang, W~Lin, IM~Navon, Y~Guo, OK~Matar,
  AG~Robins, and CC~Pain.
\newblock A reduced order model for turbulent flows in the urban environment
  using machine learning.
\newblock {\em Building and Environment}, 148:323--337, 2019.

\bibitem{xu2015data}
Tianfang Xu and Albert~J Valocchi.
\newblock Data-driven methods to improve baseflow prediction of a regional
  groundwater model.
\newblock {\em Computers \& Geosciences}, 85:124--136, 2015.

\bibitem{yang2020modified}
Zifeng Yang, Zhiqi Zeng, Ke~Wang, Sook-San Wong, Wenhua Liang, Mark Zanin, Peng
  Liu, Xudong Cao, Zhongqiang Gao, Zhitong Mai, et~al.
\newblock Modified seir and ai prediction of the epidemics trend of covid-19 in
  china under public health interventions.
\newblock {\em Journal of thoracic disease}, 12(3):165, 2020.

\bibitem{yao2018tensormol}
Kun Yao, John~E Herr, David~W Toth, Ryker Mckintyre, and John Parkhill.
\newblock The tensormol-0.1 model chemistry: a neural network augmented with
  long-range physics.
\newblock {\em Chemical science}, 9(8):2261--2269, 2018.

\bibitem{yazdani2020systems}
Alireza Yazdani, Lu~Lu, Maziar Raissi, and George~Em Karniadakis.
\newblock Systems biology informed deep learning for inferring parameters and
  hidden dynamics.
\newblock {\em PLoS computational biology}, 16(11):e1007575, 2020.

\bibitem{yildiz2019ode2vae}
Cagatay Yildiz, Markus Heinonen, and Harri L{\"a}hdesm{\"a}ki.
\newblock Ode2vae: Deep generative second order odes with bayesian neural
  networks.
\newblock In {\em Proceedings of the 33rd Conference on Neural Information
  Processing Systems (NeurIPS 2019)}, 2019.

\bibitem{zepeda2019deep}
Leonardo Zepeda-N{\'u}{\~n}ez, Yixiao Chen, Jiefu Zhang, Weile Jia, Linfeng
  Zhang, and Lin Lin.
\newblock Deep density: circumventing the kohn-sham equations via symmetry
  preserving neural networks.
\newblock {\em arXiv preprint arXiv:1912.00775}, 2019.

\bibitem{zhang2018advances}
Cheng Zhang, Judith B{\"u}tepage, Hedvig Kjellstr{\"o}m, and Stephan Mandt.
\newblock Advances in variational inference.
\newblock {\em IEEE transactions on pattern analysis and machine intelligence},
  41(8):2008--2026, 2018.

\bibitem{zhang2018real}
Liang Zhang, Gang Wang, and Georgios~B Giannakis.
\newblock Real-time power system state estimation via deep unrolled neural
  networks.
\newblock In {\em 2018 IEEE Global Conference on Signal and Information
  Processing (GlobalSIP)}, pages 907--911. IEEE, 2018.

\bibitem{zhong2019symplectic}
Yaofeng~Desmond Zhong, Biswadip Dey, and Amit Chakraborty.
\newblock Symplectic ode-net: Learning hamiltonian dynamics with control.
\newblock In {\em International Conference on Learning Representations}, 2019.

\end{thebibliography}

\end{document}